\documentclass[journal]{IEEEtran} 

\IEEEoverridecommandlockouts                             

\usepackage{CJK}
\usepackage{pifont}
\usepackage{graphicx}
\usepackage{multicol}
\usepackage{booktabs}
\usepackage{amsmath,amssymb}
\usepackage{amsthm}
\usepackage[utf8]{inputenc}
\usepackage[english]{babel}
\usepackage{multirow}
\usepackage[font=small]{caption}
\usepackage{float}
\usepackage[hidelinks]{hyperref}

\usepackage{lettrine}
\usepackage[]{algorithm2e}
\usepackage{algpseudocode}
\usepackage{cite}
\usepackage{filecontents}
\usepackage{lipsum}
\usepackage{color}
\usepackage{esdiff}
\usepackage{epstopdf}
\usepackage[normalem]{ulem}
\usepackage{soul}

\newcommand{\tabincell}[2]{\begin{tabular}{@{}#1@{}}#2\end{tabular}}
\usepackage{subcaption}
\usepackage{xcolor}
\usepackage{colortbl}
\usepackage{balance}
\newcommand{\comm}[1]{}

\definecolor{pink}{rgb}{1, 0, 1}
\definecolor{orange}{rgb}{1, 0.7529, 0}
\definecolor{darkgreen}{rgb}{0, 0.8, 0}
\begin{document}

\title{Learning-Based Motion Planning for Dynamic Environments: From Foundational Algorithms to Emerging Paradigms}

\author{
{Zongyuan Shen$^1$}, {Shalabh Gupta$^2$}, {Shancheng Zhao$^1$}, {Dehua Zhou$^1$}, {Gao Wang$^1$}, {Rui Cheng$^3$},\\{Yaming Ou$^4$}, {Zhongqiang Ren$^5$}, {Yikui Zhai$^6$}, and {C. L. Philip Chen$^7$, \textit{Life Fellow, IEEE}}
\vspace{-9pt}

\thanks {$^1$College of Information Science and Technology, Jinan University, Guangzhou 510632, China.}
\thanks {$^2$Department of Electrical and Computer Engineering, University of Connecticut, Storrs, CT 06269, USA.}
\thanks {$^3$Guangzhou Maritime University, Guangzhou 510725, China.}
\thanks {$^4$School of Artificial Intelligence, University of Chinese Academy of Sciences, Beijing 100049, China.}
\thanks {$^5$Global College, Shanghai Jiao Tong University, Shanghai 200240, China.}
\thanks {$^6$School of Electronics and Information Engineering, Wuyi University, Jiangmen 529000, China.}
\thanks {$^7$School of Computer Science and Engineering, South China University of Technology, Guangzhou 510006, China.}
}

\maketitle

\thispagestyle{empty}

\begin{abstract}

Motion planning in dynamic environments is a fundamental problem in robotics, aiming to generate safe and efficient paths, trajectories, or control actions in the presence of moving obstacles, uncertain predictions, and multi-agent interactions. It has broad applications in autonomous driving, service robotics, warehouse logistics, human-robot collaboration, crowd navigation, and multi-robot systems. This survey reviews representative works published primarily between 2015 and 2025, with a particular focus on how recent learning-based advances extend, complement, or interact with classical planning foundations. We first revisit classical planning methods as algorithmic foundations and reference frameworks for learning-based extensions. We then propose a role-of-learning taxonomy that categorizes existing methods according to how learning participates in the planning pipeline, including direct policy learning, learning-augmented classical planning, hybrid planning, and training enhancement methods. For each category, we summarize the main problem settings, representative algorithms, key ideas, integration mechanisms, strengths, and limitations. We further analyze how observation representations, prediction uncertainty, interaction modeling, planner integration, safety constraints, and training strategies shape learning-based motion planning in dynamic environments. Finally, we discuss open challenges and future directions, including sim-to-real gap, safe and certifiable planning, dense crowd navigation, perception-planning coupling, and embodied AI.
\end{abstract}

\begin{IEEEkeywords}
Motion planning, reinforcement learning, dynamic environments, collision avoidance, autonomous robots.

\vspace{-8pt}

\end{IEEEkeywords}
\section{Introduction}

\begin{figure}[t]
\vspace{4pt}
    \centering
    \subfloat[Social-aware navigation~\cite{chen2017decentralized}]{
    \includegraphics[width=0.23\textwidth]{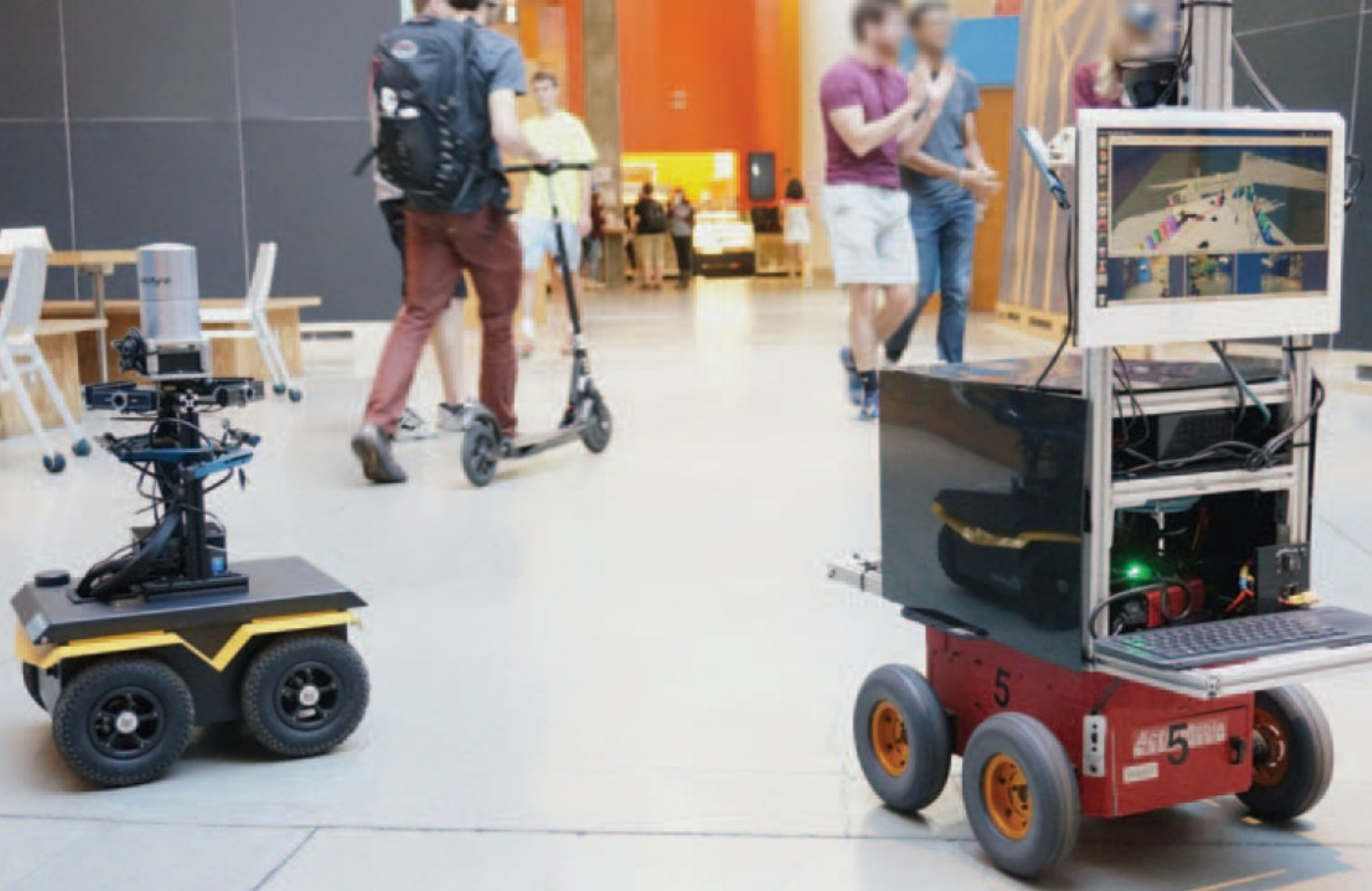}\label{fig:example1}}\hspace{-10pt}\quad
    \centering
    \subfloat[Robotic manipulation~\cite{Liu2025_flexible}]{
    \includegraphics[width=0.23\textwidth]{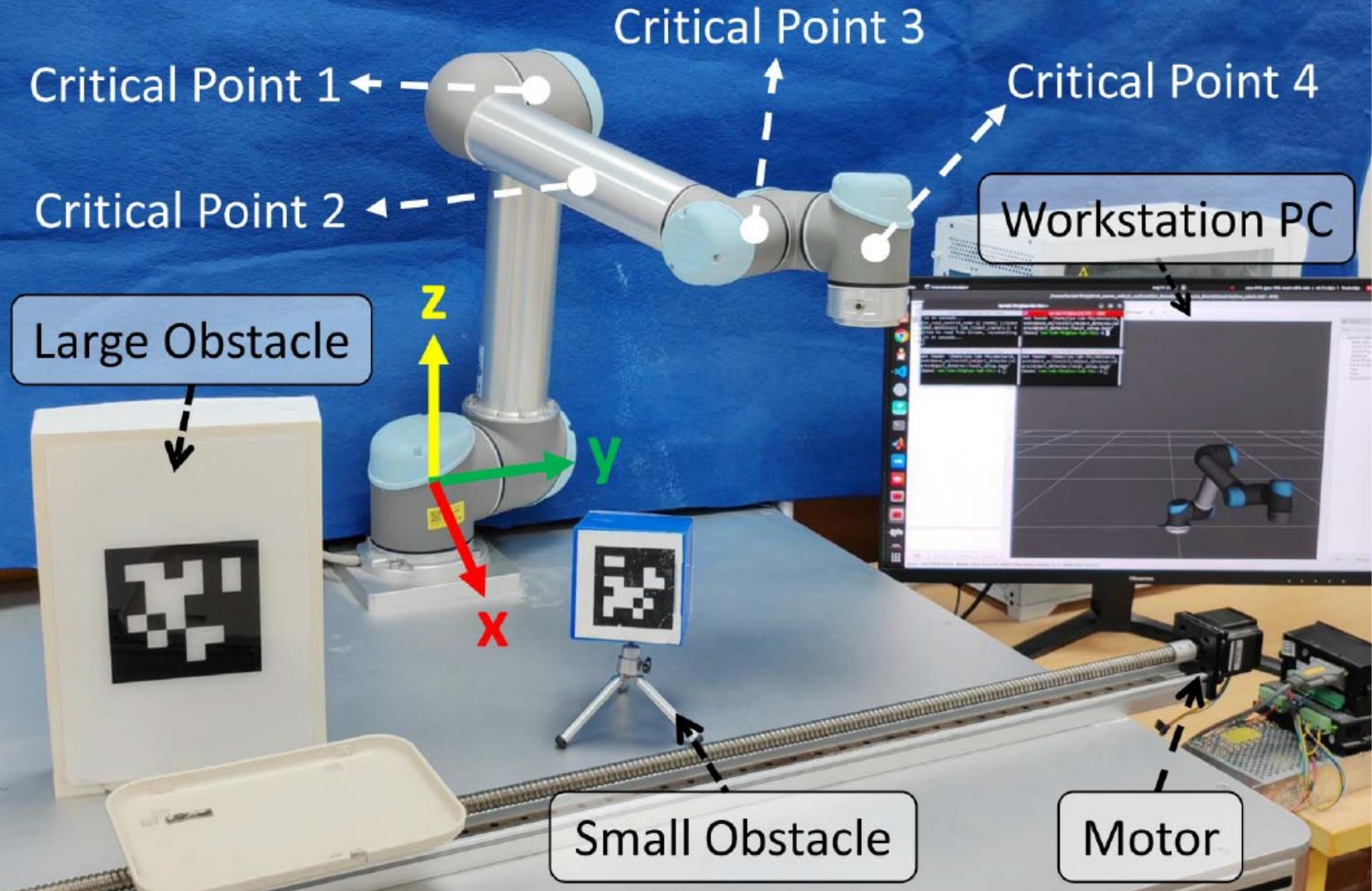}\label{fig:example2}}\vspace{0.5em}\\
    \centering
    \subfloat[Multi-agent navigation~\cite{Kondo2024}]{
    \includegraphics[width=0.23\textwidth]{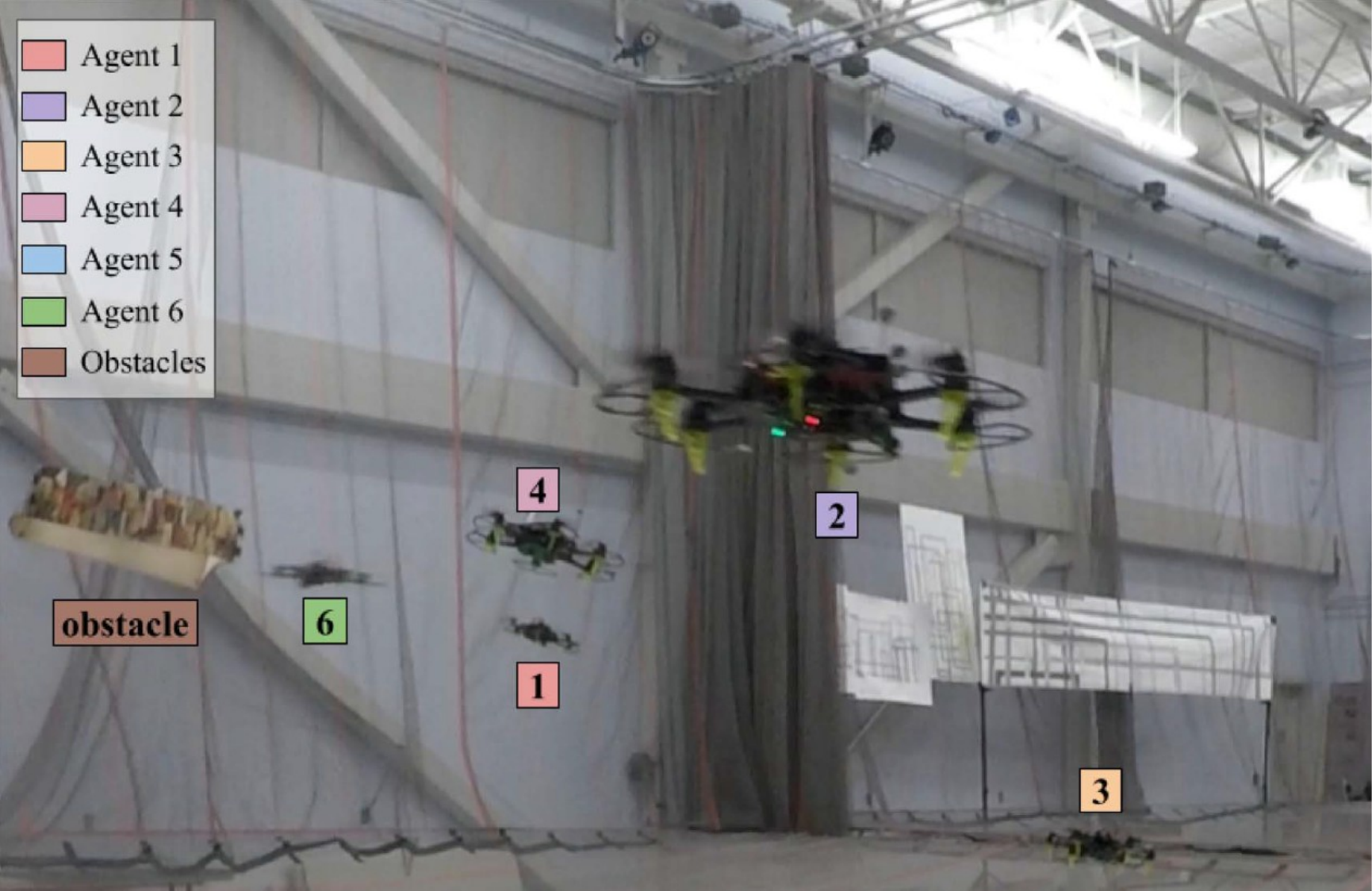}\label{fig:example3}}\hspace{-10pt}\quad
    \centering
    \subfloat[Autonomous driving~\cite{Sun2021_sparse}]{
    \includegraphics[width=0.23\textwidth]{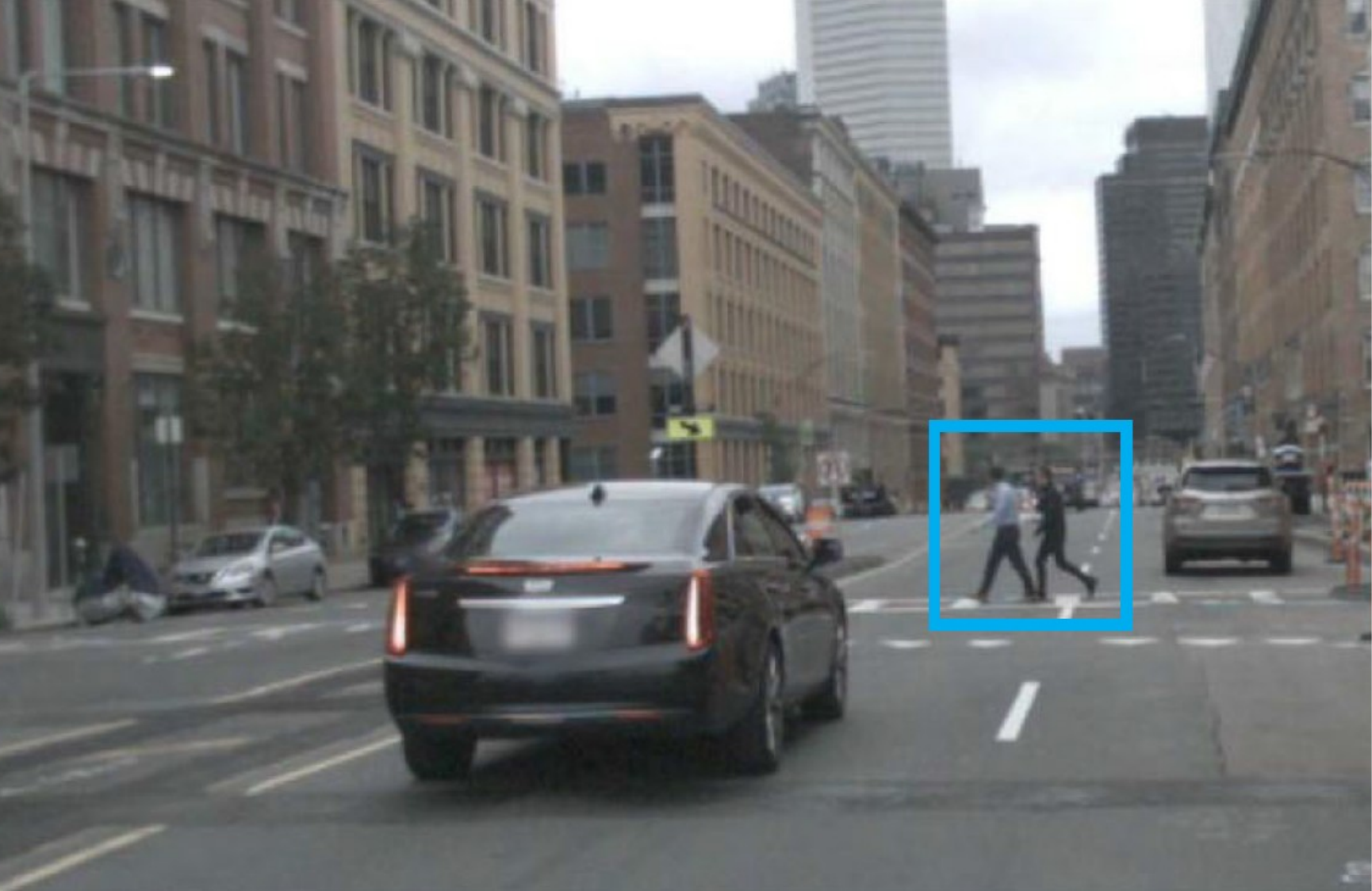}\label{fig:example4}}\\
          \caption{Application examples of motion planning in dynamic environments: (a) mobile robot navigation in public spaces~\cite{chen2017decentralized}, (b) robotic manipulation in dynamic workspaces~\cite{Liu2025_flexible}, (c) aerial navigation among agents~\cite{Kondo2024}, and (d) autonomous driving with moving humans~\cite{Sun2021_sparse}.}\label{fig:example}
          \vspace{-1em}
\end{figure}

Autonomous robots are increasingly deployed in dynamic environments to perform different activities, such as public service, warehouse operations, and human-robot collaborative tasking. In these scenarios, robots must navigate among pedestrians, vehicles, other robots, and moving objects whose future trajectories may be uncertain or difficult to predict~\cite{hare2020pose}. Unlike motion planning in static environments, dynamic environments require continuous plan adaptation rather than one-time path generation. The robot must update its trajectory or control actions in real time to avoid collisions, maintain steady progress, and ensure safe and uninterrupted navigation. Fig.~\ref{fig:example} shows application examples of motion planning in dynamic environments using different robotic platforms, such as unmanned  ground vehicles, industrial manipulators, unmanned aerial vehicles, and driverless cars.

\subsection{Motivation}
The above requirements make motion planning in dynamic environments challenging due to a constant change in the feasible navigation space of the robot as obstacles move, appear, or disappear. Furthermore, the motion of pedestrians, vehicles, and other robots is often uncertain, making it difficult to predict future collision risks~\cite{shen2023smart}. In particular, in human-populated or multi-agent scenarios, the robot must also account for robot-agent and agent-agent interactions, since the motion of each agent may influence the decisions of others~\cite{jiang2024learning,Lu2025,Zhou2025}. Moreover, the planning decisions must be updated in real time while satisfying collision-avoidance, kinodynamic constraints, and task-progress requirements. Failure to balance these requirements may lead to unsafe motions, oscillatory behaviors, excessive conservatism, or the freezing robot problem.

Classical planning methods~\cite{otte2016rrtx,koenig2004lifelong,fiorini1998motion,ge2002dynamic} have long provided the foundation for motion planning in dynamic environments. These methods can be grouped into global and local replanning methods. Global replanning methods target optimality and provide long-range path guidance to produce feasible paths when environmental changes invalidate the current path, whereas local replanning methods repair robot trajectories over a short-horizon or generate control commands for real-time collision avoidance. The classical methods are widely used because they provide interpretable decision rules and explicit mechanisms for safety or feasibility with reasoning. However, their performance often depends on hand-crafted costs, manually tuned parameters, simplified interaction models, reliable obstacle prediction, and sufficient online computational resources. 

To overcome these limitations, learning-based methods have been increasingly applied to motion planning in dynamic environments. By learning from demonstrations, interactions, or simulated experiments, these methods can build navigation policies, interaction patterns, adaptive costs, planning guidance, and dynamic-obstacle predictions from data. Importantly, learning is not limited to replacing planners with end-to-end policies; it can also enhance classical planners or be integrated with them in hybrid planners. In direct policy learning~\cite{long2017deep,xie2021towards,xie2023drl}, the learned models serve as the primary decision-making module, whereas in learning-enhanced~\cite{Zhu2025,Han2025,dobrevski2024dynamic} and hybrid planning~\cite{Xu2024,fan2020distributed,hong2023obstacle}, the learned modules provide auxiliary guidance, intermediate decisions, or online adaptation while classical planners retain the primary structural planning roles. This diversity makes it necessary to review learning-based motion planning methods according to how the learning approaches contribute to the planning process.

\begin{figure}[t]
   \centering        
    \includegraphics[width=0.50\textwidth]{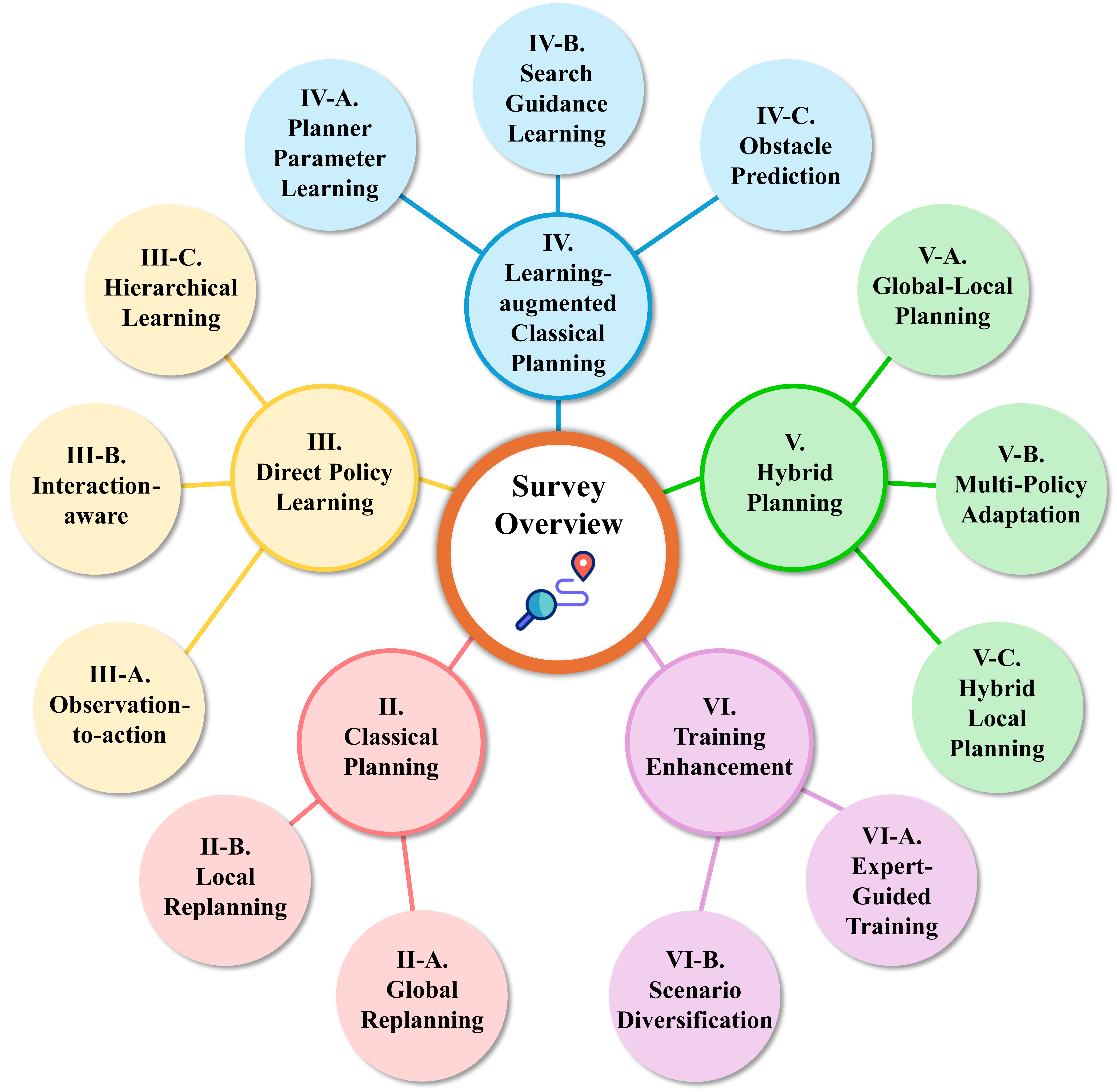} 
    \caption{Taxonomy of motion planning in dynamic environments.}\label{fig:surveyoverview} 
    \vspace{-1em}
 \end{figure}

\subsection{Existing Surveys}
Although motion planning in static environments has been extensively reviewed~\cite{kingston2018sampling,gammell2021asymptotically,orthey2023sampling,liu2023path}, surveys specifically dedicated to motion planning in dynamic environments remain relatively limited. The survey by Kamil et al.~\cite{kamil2015review} focused mainly on local replanning methods, such as potential field-based, control-based, velocity obstacle-based methods. Mohanan et al.~\cite{mohanan2018survey} provided an in-depth review of classical planning methods for dynamic environments. These surveys summarize developments up to 2015 but do not reflect the significant progress made in learning-based motion planning over the past decade. More recent surveys have focused on specific problem settings or application domains. Hewawasam et al.~\cite{hewawasam2022past} briefly reviewed path planning for mobile robot navigation in dynamic environments. Liu et al.~\cite{liu2024review} provided a focused review on motion planning for industrial manipulators operating in dynamic environments, with particular attention to real-time planning performance in high-dimensional configuration spaces.

\subsection{Contributions}
Although existing surveys provide valuable insights, a unified and up-to-date review of learning-based motion planning in dynamic environments is still lacking. This survey addresses this gap by presenting a comprehensive and structured review of representative works primarily published between 2015 and 2025, while linking them to classical planning foundations. Specifically, before reviewing learning-based motion planning methods, we first revisit classical planning methods as algorithmic foundations and reference frameworks for later learning-based extensions. Rather than organizing the literature only by learning paradigms or classical algorithmic families, we adopt a role-of-learning taxonomy that characterizes how learning participates in the planning pipeline. 

Accordingly, existing learning-based motion planning methods are organized into four main categories: direct policy learning, learning-augmented classical planning, hybrid planning, and training enhancement. The roles and key distinctions of these categories are summarized in Table~\ref{tab:methodtaxonomy}. For each category, we review the main problem settings, representative algorithms, key ideas, learning paradigms, integration mechanisms, and strengths and limitations. Fig.~\ref{fig:surveyoverview} shows overall organization of these categories and the methodologies therein.

\begin{table*}[t]
\centering
\caption{Taxonomy and comparison of learning-based motion planning methods in dynamic environments.}\vspace{-3pt}
\label{tab:methodtaxonomy}
\centering
\setlength\tabcolsep{5pt}
\begin{tabular}{l l l l l l}
\toprule
\specialrule{0.1em}{1pt}{1pt} 
\tabincell{l}{\textbf{Category}} 
&\tabincell{l}{\textbf{Main Idea}} 
&\tabincell{l}{\textbf{Role of Learning}} 
&\tabincell{l}{\textbf{Generated Output}}
&\tabincell{l}{\textbf{{Strengths}}}
&\tabincell{l}{\textbf{Limitations}}\\
\toprule

\tabincell{l}{\textbf{Direct Policy}\\\textbf{Learning}} 
&\tabincell{l}{{Learn navigation}\\{policies directly}\\{from observations}} 
&\tabincell{l}{{Primary navigation}\\{decision-making}\\{module}} 
&\tabincell{l}{{Navigation actions,}\\{local trajectories,}\\{and subgoals}} 
&\tabincell{l}{{Fast and compact}\\{end-to-end navigation}\\{decision making}}
&\tabincell{l}{{Limited generalization,}\\{interpretability, and}\\{safety guarantees}}\vspace{0.5em}\\

\specialrule{0em}{2pt}{2pt}
\tabincell{l}{\textbf{Learning-}\\\textbf{Augmented}\\\textbf{Classical}\\\textbf{Planning}} 
&\tabincell{l}{{Use learned module}\\{to enhance, rather than}\\{replace, classical planners}} 
&\tabincell{l}{{Auxiliary module}} 
&\tabincell{l}{{Planner parameters,}\\{search guidance, and}\\{obstacle predictions}} 
&\tabincell{l}{{Improve adaptability}\\{in complex scenarios,}\\{maintain interpretability}\\{of classical planners}}
&\tabincell{l}{{Sensitive to the quality}\\{and generalization of}\\{learned auxiliary}\\{information}}\vspace{0.5em}\\

\specialrule{0em}{2pt}{2pt}
\tabincell{l}{\textbf{Hybrid}\\\textbf{Planning}} 
&\tabincell{l}{{Couple learned decision}\\{module with classical}\\{planners in the online}\\{planning pipeline}} 
&\tabincell{l}{{Navigation decision}\\{making module coupled}\\{with classical planner}}
&\tabincell{l}{{Navigation actions,}\\{policy switching}\\{decisions, and subgoals}} 
&\tabincell{l}{{Balances learning}\\{adaptability with safety}\\{and interpretability}\\{of classical planners}}
&\tabincell{l}{{Sensitive to module}\\{coupling, learned-}\\{output quality, and}\\{policy generalization}}\vspace{0.5em}\\

\specialrule{0em}{2pt}{2pt}
\tabincell{l}{\textbf{Training}\\\textbf{Enhancement}} 
&\tabincell{l}{{Improves policy learning}\\{through expert guidance}\\{or diversified scenarios}} 
&\tabincell{l}{{Policy improvement}\\{at training stage}} 
&\tabincell{l}{{Expert-guided initial}\\{policies, and diverse}\\{training scenarios}} 
&\tabincell{l}{{Improves training}\\{efficiency, robustness,}\\{and generalization}}
&\tabincell{l}{{Performance depends}\\{on expert quality,}\\{training diversity, and}\\{scenario coverage}}\\
\bottomrule
\specialrule{0.1em}{1pt}{1pt} 
\end{tabular}
\vspace{-1.0em}
\end{table*}

\subsection{Organization}
The remainder of this paper is organized as follows. Section~\ref{sec:classicalplanner} reviews classical planning foundations for dynamic environments, including global replanning and local replanning methods. Section~\ref{sec:directpolicylearning} discusses direct-policy learning methods, where learned models serve as the primary navigation decision maker. Section~\ref{sec:learningAugmented} reviews learning-enhanced classical planning methods, in which learning is used to improve classical planners through learned parameter, search guidance, or obstacle prediction. Section~\ref{sec:hybridplanning} discusses hybrid planning methods that couple learned modules with classical planners in the online planning pipeline. Section~\ref{sec:learningenhancement} reviews training enhancement methods that improve policy learning through expert guidance or diverse scenarios. Finally, Section~\ref{sec:conclusions} concludes the survey and discusses future research directions.
\section{Classical Planning Foundations}
\label{sec:classicalplanner}

Classical planning methods provide the algorithmic foundations for navigation in dynamic environments. They define core mechanisms that are later enhanced or integrated with learning-based modules, including sampling, graph search, velocity selection, potential-field design, and control command generation. This section groups the representative classical methods into two types: global and local replanning methods.
 
\subsection{Global Replanning Methods}

Global replanning methods provide long-range planning by updating and repairing the robot trajectory when dynamic obstacles or environmental changes invalidate it or pose a risk. In this category, the existing methods can be classified into sampling-based and search-based methods.

\subsubsection{Sampling-Based Methods}

Sampling-based methods typically extend Probabilistic Roadmap (PRM)~\cite{kavraki1996probabilistic}, Rapidly-exploring Random Tree (RRT)~\cite{lavalle2001randomized}, and RRT*~\cite{karaman2011sampling} to dynamic environments. They can be broadly classified as reactive or active. Reactive methods rely on current observations to update paths, enabling fast responses but lacking foresight in complex environments. Active methods incorporate predicted future states of dynamic obstacles for risk-aware replanning. However, their performance depends on prediction accuracy and may degrade under limited sensing or dense crowds.

a) \textit{Reactive Methods:} Early reactive methods reused previous search efforts to some extent to accelerate replanning. Bruce and Veloso~\cite{bruce2002real} rebuilt the tree from scratch while biasing new samples toward the previous path. Ferguson et al.~\cite{ferguson2006replanning} improved tree reuse by pruning invalidated nodes and their descendants to retain a single-tree structure. Zucker et al.~\cite{zucker2007multipartite} further preserved valid tree components by pruning only colliding nodes and reconnecting the resulting disjoint trees through biased sampling. To fully reuse previous search efforts, Otte and Frazzoli~\cite{otte2016rrtx} maintained the same graph without pruning nodes and repaired the goal-rooted subtree through rewiring cascades after environmental changes. Later variants improve convergence, reduce computation, or extend the idea to kinodynamic replanning~\cite{huang2023fast,huang2024asymptotically,silveira2023real}.

More recent methods develop advanced heuristics to guide tree repair. Chen et al.~\cite{chen2019horizon} biased new samples toward low-cost regions captured by a Gaussian mixture model trained online. Other methods use dual-tree, reverse-tree, potential-field, or multi-objective guidance to accelerate reconnection and improve path quality~\cite{yuan2020efficient,qi2021,lee2023path,cui2024rt}. Shen et al.~\cite{shen2023smart} performed fast tree-repair at hot spots that lie at the intersection of different disjoint trees, reducing unnecessary random exploration and improving computational efficiency.

b) \textit{Active Methods:} Early methods mainly used prediction-based risk to regulate node insertion and pruning. Chiang et al.~\cite{chiang2017dynamic} gradually increased risk tolerance over time to account for growing prediction uncertainty. Chi et al.~\cite{chi2019risk} expanded low-risk nodes in a time-embedded tree within a predictive horizon and then improved path quality through rewiring.

Later methods improved active replanning through enhanced risk modeling and search structures. Cai et al.~\cite{cai2023human} expanded the tree by adding low-risk nodes, where the risk is evaluated based on collision probability and the likelihood of entering dense human crowds. Hakobyan and Yang~\cite{hakobyan2023distributionally} quantified node risk with a safety loss based on conditional value-at-risk, which assigns higher risk to nodes that may lead to severe safety violations. This allows the planner to distinguish nodes with similar collision probabilities but different severities of potential safety violations. Sun et al.~\cite{sun2024multi} expanded multiple trees to explore the environment in parallel and used them to guide the main tree toward safer regions, thereby improving search efficiency and robustness in cluttered environments.

\subsubsection{Search-Based Methods} Search-based methods provide another classical foundation for global replanning. They represent the configuration space as a graph, grid, or lattice, and utilize fundamental graph-search algorithms, such as Dijkstra's algorithm~\cite{dijkstra1959note} and A*~\cite{hart1968formal}. Subsequently, these methods employ techniques such as incremental graph repair, spatio-temporal safety reasoning, and kinodynamic primitive search.

a) \textit{Incremental Graph Repair:} These methods improve replanning efficiency by updating only the affected parts of the graph when environmental changes modify edge costs. Early methods, such as D*~\cite{stentz1995focussed}, D* Lite~\cite{koenig2002d}, and Lifelong Planning A* (LPA*)~\cite{koenig2004lifelong}, repair A*-like solutions according to newly acquired environmental information. Ren et al.~\cite{ren2022multi} extended this line of work to multi-objective replanning by maintaining a Pareto-optimal set of paths under vector-valued edge costs. Li et al.~\cite{li2023bidirectional} combined forward and backward searches to improve the reuse of previous search results during replanning.

b) \textit{Spatio-Temporal Safety Reasoning:} These methods search for collision-free paths in the spatio-temporal domain. Safe Interval Path Planning~\cite{phillips2011sipp} associates each configuration with collision-free time intervals and searches over these intervals rather than every discrete time step, thereby reducing the search space while preserving completeness and optimality under its assumptions. Several variants improve its efficiency and applicability~\cite{phillips2011planning,narayanan2012anytime,gonzalez2012using,ren2022MO}. Another line of work builds time-varying graphs. Cao et al.~\cite{cao2019dynamic} constructed a graph from the Delaunay triangulation of pedestrian positions, where edges define time-varying gates between neighboring pedestrians. An A*-based search is then used to find a feasible channel through these gates. Huang et al.~\cite{huang2025safe} constructed a dynamic connected visibility graph with safe intervals on graph edges, from which spatially and temporally distinct initial paths and safe corridors are generated for trajectory optimization.

c) \textit{Kinodynamic Primitive Search:} These methods extend graph search from geometric path planning to kinodynamic trajectory generation by expanding short-horizon motion primitives that satisfy robot motion constraints while avoiding dynamic obstacles. Lin et al.~\cite{lin2021search} proposed a search-based online partial motion planner for car-like robots, which generates motion primitives by discretizing time and control inputs and searches a state-time graph for feasible motion. 

Later methods incorporate prediction, risk modeling, and hierarchical design. Chen et al.~\cite{chen2022rast} used a particle-based dynamic map to predict future occupancy and an A*-based search to generate a reference path, which is converted into spatio-temporal safety corridors for trajectory optimization. Chen et al.~\cite{chen2023risk} combined local risk-aware primitive search with global reference guidance, enabling the robot to follow the global path when safe, deviate locally when blocked, and reconnect once a safe path becomes available. Qi et al.~\cite{qi2023hierarchical} generated a coarse trajectory through spatio-temporal motion-primitive search and refined the path and speed profile via multilayer optimization. Wiman and Tiger~\cite{wiman2025safe} used a high-resolution lattice for short-term collision-free planning and a low-resolution lattice for long-horizon guidance.

\textit{Motivation for Learning-Based Planning Methods:} Classical global replanning methods support long-range path generation and online path repair, but their performance is often constrained by sampling efficiency, graph resolution, heuristic design, and prediction accuracy. These limitations motivate learning-based planning methods that improve global replanning by learning sampling biases, search costs, heuristic functions, risk maps, or obstacle predictions. In hybrid frameworks, classical global replanners provide long-range guidance, while learned local modules handle collision avoidance, kinodynamic constraints, or interaction-aware decision making.

\subsection{Local Replanning Methods}

Local replanning methods provide short-range planning by updating and repairing the robot trajectory when dynamic obstacles or environmental changes invalidate it or pose a risk. These methods generate control commands for a short time-horizon based on the local information. Existing methods are grouped into model predictive control (MPC), velocity obstacle, potential field, and dynamic window methods.

\subsubsection{Model Predictive Control Methods} 

MPC methods formulate local replanning as a finite-horizon optimal control problem. At each planning step, MPC predicts the robot's future states over a finite horizon and incorporates predicted obstacle motion into safety constraints. It then optimizes control inputs under robot dynamics and safety constraints. 

A central concern is how to formulate safety constraints for collision avoidance using predicted obstacle trajectories and uncertainties. Existing methods commonly represent obstacles as convex polyhedra (e.g., cuboids) or smooth differentiable surfaces (e.g., ellipsoids)~\cite{castillo2020real}. Polyhedral models can encode collision avoidance via linear inequality constraints, but can be computationally expensive in crowded scenarios. Smooth surface representations reduce the number of constraints and are better suited to nonlinear optimization, but may yield conservative solutions. Based on these representations, collision avoidance can be formulated as deterministic constraints~\cite{Brito2019}, chance constraints~\cite{Blackmore2011,Zhu2019,castillo2020real}, scenario constraints~\cite{Groot2021}, or control barrier function-based constraints~\cite{jian2023dynamic,huang2025risk,saviolo2025reactive}.

Another concern is deadlock caused by local minima in the optimization landscape. Existing methods mainly address this through high-level guidance or proactive cost design. Groot et al.~\cite{Groot2025} generated homotopically distinct global paths using a visibility-probabilistic roadmap~\cite{simeon2000visibility}, tracked them with parallel MPC planners, and selects the MPC trajectory with the lowest cost. Arul et al.~\cite{arul2023ds} introduced a terminal state cost based on expected time-to-goal and time-to-collision, which favors the states with better goal-reaching potential and more free space ahead. This encourages the robot to detour rather than freeze in locally safe but deadlocked configurations.

\subsubsection{Velocity Obstacle Methods}

Velocity obstacle methods perform local collision avoidance in velocity space. The concept of Velocity Obstacles (VO), introduced by Fiorini and Shiller~\cite{fiorini1998motion}, characterizes the set of relative velocities that would lead to future collisions between the robot and nearby agents. By selecting a velocity outside the VO set at each planning step, the robot can maintain collision-free motion.

Oscillations may occur when agents simultaneously react to each other's previous velocity choices. To address this issue, Van den Berg et al.~\cite{van2008reciprocal} proposed Reciprocal Velocity Obstacles (RVO), which assumes that both agents share collision-avoidance effort when constructing the VO set, thereby reducing overreactions. Snape et al.~\cite{snape2011hybrid} introduced Hybrid RVO (HRVO) to further reduce oscillatory behaviors. 

More recent methods relax the ideal cooperation assumptions. Liu et al.~\cite{LiuZhihao2024} used reinforcement learning (RL) to infer collision-avoidance responsibility, which is the relative share of the collision avoidance effort that the robot should take when interacting with nearby agents. Martinez et al.~\cite{Martinez2025} used an adaptive control law to estimate the cooperation level, which describes how much those agents are expected to participate in the reciprocal collision avoidance of nearby agents, and accordingly adjust the robot's responsibility. 

Other works extend VO method from holonomic models to more realistic motion models by incorporating smoothness requirements~\cite{snape2010smooth}, acceleration limits~\cite{van2011reciprocal}, linear dynamics constraints~\cite{bareiss2013reciprocal}, and nonholonomic motion constraints~\cite{zhao2022solving}.

\subsubsection{Potential Field Methods}

Potential field methods define attractive potentials toward the goal and repulsive potentials around obstacles, guiding the robot by the negative gradient of the combined field. For dynamic environments, Ge and Cui~\cite{ge2002dynamic} constructed potentials from relative position and velocity between the robot, goal, and obstacles. Later methods improved dynamic avoidance using global sampling guidance~\cite{chiang2015path}, stochastic reachable sets for uncertain obstacles~\cite{malone2017hybrid}, oval or bounded vector fields for smoother bypassing and goal convergence~\cite{boldrer2020socially,huber2022avoiding}, and vortex-based or emergency deflection mechanisms to escape local minima~\cite{wang2026enhanced}.

\subsubsection{Dynamic Window Methods}

Dynamic Window Approach (DWA)~\cite{fox1997dynamic} performs local replanning by evaluating the candidate control commands under the robot's dynamic constraints. Given the velocity and acceleration limits of the robot, DWA constructs a dynamic window of admissible commands, scores each candidate according to certain criteria (e.g., goal progress and obstacle clearance), and executes the best command. Extensions for dynamic environments include dynamic-polygon obstacle representations~\cite{missura2019predictive}, long-term DWA with Elastic Band refinement~\cite{jian2023long,quinlan1993elastic}, uncertainty-aware safe DWA~\cite{yasuda2023safe}, and gradient-aware collision costs based on obstacle-distance fields~\cite{zhang2025gradient}.

\textit{Motivation for Learning-Based Planning Methods:} Classical local replanning methods provide efficient mechanisms for real-time collision avoidance, but their performance often depends on hand-crafted cost functions, manually tuned parameters, interaction assumptions, and reliable short-term perception or prediction. These limitations motivate the learning-based methods that improve local replanning by learning adaptive costs, planner parameters, obstacle predictions, or interaction-aware behaviors. Learning can either enhance classical planners with auxiliary guidance or be integrated with them in hybrid frameworks to combine data-driven decision making with model-based safety and feasibility.

\begin{table*}[t]{}
\footnotesize
\caption {Comparison of representative observation-to-action methods.}\label{tab:obs_to_action_table}\vspace{-3pt}
\centering
\setlength\tabcolsep{6pt}
\begin{tabular}{l l l l l l l} 
 \toprule
\specialrule{0.1em}{1pt}{1pt} 
\tabincell{c}{\textbf{Category}}
&\multicolumn{1}{l}{\textbf{Reference}}
&\tabincell{c}{\textbf{Year}}
&\tabincell{c}{\textbf{Main Idea}}
&\tabincell{c}{\textbf{Sensing Modality}}
&\tabincell{c}{{\textbf{Learning Paradigm}}}  
&\tabincell{c}{\textbf{Policy Output}}\\ 
\toprule

\multirow{9}{*}{\tabincell{l}{\textbf{{Single-Modal}}}}
& \tabincell{l}{\cite{long2017deep}}       & 2017     &\tabincell{l}{{Discretizes the velocity space}\\{and learns a multi-class classifier}\\{from ORCA-generated data for}\\{collision-free velocity selection}} & \tabincell{l}{LiDAR}  &\tabincell{l}{{Supervised learning}} &\tabincell{l}{{Control command}}  \vspace{0.5em}\\

& \tabincell{l}{\cite{long2018towards}}      & 2018     &\tabincell{l}{{Trains a shared policy using}\\{PPO for decentralized multi-}\\{robot collision avoidance}} & \tabincell{l}{LiDAR}  &\tabincell{l}{{Reinforcement learning}} &\tabincell{l}{{Control command}}  \vspace{0.5em}\\

& \tabincell{l}{\cite{Heuvel2024}}      & 2024     &\tabincell{l}{{Uses a DDPG-based policy}\\{with spatio-temporal attention}\\{to capture obstacle motion}\\{trends from LiDAR observations}} & \tabincell{l}{LiDAR}  &\tabincell{l}{{Reinforcement learning}} &\tabincell{l}{{Control command}}  \\

\specialrule{0.05em}{2pt}{2pt}
\multirow{14}{*}{\tabincell{l}{\textbf{{Multi-Modal}}}}
& \tabincell{l}{\cite{xie2021towards}}      & 2021     &\tabincell{l}{{Trains a CNN-based policy as}\\{a continuous velocity regressor}\\{from DWA-generated data for}\\{control command generation}} & \tabincell{l}{{LiDAR + Camera}}  &\tabincell{l}{{Supervised learning}} &\tabincell{l}{{Control command}}  \vspace{0.5em}\\

& \tabincell{l}{\cite{huang2021towards}}       & 2021     &\tabincell{l}{{Fuses semantic segmentation}\\{maps and LiDAR data into a}\\{PPO-based policy to improve}\\{navigation robustness}} & \tabincell{l}{{LiDAR + Camera}}  &\tabincell{l}{{Reinforcement learning}} &\tabincell{l}{{Control command}}  \vspace{0.5em}\\

& \tabincell{l}{\cite{han2022deep}}      & 2022     &\tabincell{l}{{Weights depth features using}\\{self-state attention and feeds}\\{them into a PPO-based policy}} & \tabincell{l}{{LiDAR + Camera}}  &\tabincell{l}{{Reinforcement learning}} &\tabincell{l}{{Control command}}  \vspace{0.5em}\\

& \tabincell{l}{\cite{xie2023drl}}      & 2023     &\tabincell{l}{{Uses a VO-based reward to}\\{train a PPO-based policy for}\\{proactive collision avoidance}} & \tabincell{l}{{LiDAR + Camera}}  &\tabincell{l}{{Reinforcement learning}} &\tabincell{l}{{Control command}}\\

\bottomrule
\specialrule{0.1em}{1pt}{1pt}%\vspace{2pt}
\end{tabular}
\vspace{-1.0em}
\end{table*}

\section{Direct Policy Learning Methods}
\label{sec:directpolicylearning}

In learning-based motion planning, learning generally refers to using data or interaction experience to train parameterized models that support navigation decision making. In robotic applications, these models are commonly implemented as neural networks, such as Multi-Layer Perceptron (MLP), Convolutional Neural Network (CNN), Recurrent Neural Network (RNN), and Graph Neural Network (GNN). Depending on the training formulation, they can be optimized by supervised learning, imitation learning, RL, or hybrid training schemes. The learned models may serve different roles, such as extracting observation features, encoding agent interactions, predicting future states, or generating navigation actions.

Direct policy learning methods use learned models in the primary decision-making module for navigation in dynamic environments. Given sensor observations, robot states, agent states, or interaction models, the learned policies directly generate control commands, local trajectories, subgoals, or navigation behaviors. According to the policy formulation, existing methods can be grouped into three classes: observation-to-action, interaction-aware, and hierarchical learning methods.

\subsection{Observation-to-Action Methods}
\label{subsec:obsToAction}

Observation-to-action methods directly map robot observations to navigation actions. By reducing the reliance on explicit environmental modeling or agent-level reasoning, they provide a compact decision-making pipeline for navigation in dynamic environments. Existing methods can be classified into single-modal and multi-modal methods, depending on whether the policy relies on one sensing modality or combines complementary sensory information. Table~\ref{tab:obs_to_action_table} summarizes the key features of representative observation-to-action methods.

\subsubsection{Single-Modal Methods}

Single-modal methods learn navigation policies based on one primary sensing modality. 

a) \textit{Supervised Learning-Based Methods:} These methods learn navigation policies from labeled data. Long et al.~\cite{long2017deep} generated training data using Optimal Reciprocal Collision Avoidance (ORCA)~\cite{Berg2011_ORCA} under different parameter settings and sensing-noise levels. The continuous velocity space is discretized, and a network is trained as a multi-class classifier to infer collision-avoidance velocities from LiDAR measurements and a goal-directed preferred velocity. 

b) \textit{Reinforcement Learning-Based Methods:} These methods learn navigation policies through reinforcement learning (RL) by improving the policy through interactions with the environment. Long et al.~\cite{long2018towards} proposed a LiDAR-based method that maps raw LiDAR measurements to control commands for decentralized multi-robot collision avoidance. The method formulates the task as a Partially Observable Markov Decision Process (POMDP)~\cite{littman2009tutorial} and trains a shared policy using Proximal Policy Optimization (PPO)~\cite{schulman2017proximal}. Fan et al.~\cite{Fan2019} extended this method to dense crowds by introducing feature-rich recovery points, which mitigate localization failures caused by crowd occlusions and degraded LiDAR-based SLAM. Heuvel et al.~\cite{Heuvel2024} further exploited temporal structure in LiDAR observations. The method constructs Temporal Accumulation Group Descriptors (TAGDs) from consecutive LiDAR measurements to capture obstacle motion trends. A Deep Deterministic Policy Gradient (DDPG)-based policy~\cite{lillicrap2015continuous} then takes LiDAR measurements, TAGDs, upcoming waypoints of robot as inputs and processes them via separate spatial and temporal attention streams, producing feature embeddings, which are concatenated and processed to generate control commands. Fan et al.~\cite{Fan_fly2025} compressed historical LiDAR scans into a 2D obstacle map that captures obstacle contours and motion cues. A PPO-trained network combines this temporal representation with the quadrotor state and target command to output horizontal acceleration commands. A dynamic obstacle-aware reward enlarges the risk region along obstacle motion directions, improving reactions to high-speed obstacles.

\textit{Strengths and Limitations:} Single-modal methods are computationally efficient and easy to implement; however, relying on a single sensing modality limits the utilization of rich environmental information. Especially in complex scenarios, a single sensing modality may not fully capture the semantic cues, spatial structure, and dynamic obstacle motion. These limitations motivate multi-modal methods that incorporate complementary sensory information.

\subsubsection{Multi-Modal Methods}

Multi-modal methods learn policies from complementary sensing modalities (e.g., camera and LiDAR) to enrich the representation of dynamic environments. 

a) \textit{Supervised Learning-Based Methods:} These methods learn navigation policies from labeled data. Pokle et al.~\cite{pokle2019deep} trained a local planner via behavioral cloning from simulated human teleoperation and the ROS Navigation Stack with social costs~\cite{lu2014layered}. Given a global path, LiDAR measurements, robot odometry, and nearby human trajectories, the method generates a local trajectory. Xie et al.~\cite{xie2021towards} used DWA~\cite{fox1997dynamic} as the expert to generate training data. The method adopts a CNN~\cite{lecun1998gradient} that takes a short history of LiDAR measurements, human states, and a waypoint as inputs, and learns continuous velocity commands by minimizing the mean-square-error loss between expert and predicted velocities.

b) \textit{Reinforcement Learning-Based Methods:} These methods learn navigation policies through RL. Huang et al.~\cite{huang2021towards} incorporated visual semantics and LiDAR measurements into an RL-based policy. The method first converts RGB images into binary traversability maps, which are then fused with LiDAR measurements and processed by a PPO-trained policy. Han et al.~\cite{han2022deep} focused on enhancing depth perception from RGB images and LiDAR measurements. The method uses sparse-to-dense depth completion~\cite{ma2018sparse} to estimate a depth map, which is fused with LiDAR data into a 2D obstacle-distance representation. A self-state-attention module assigns adaptive weights to depth features, and the resulting representation is used for policy learning. Xie and Dames~\cite{xie2023drl} combined preprocessed sensor observations with VO-based reward shaping in RL. The method converts a short history of LiDAR scans into a grid map to encode geometric structure. It detects and tracks pedestrians from RGB-D data, and encodes their relative states into pedestrian kinematic maps. These representations, together with a subgoal, are used as policy inputs. It further uses VO~\cite{fiorini1998motion} to formulate an active heading reward, which encourages the robot to steer toward a collision-free direction while maintaining progress toward the goal.

\textit{Strengths and Limitations:} Multi-modal methods provide both semantic information and geometric structure by combining visual inputs and LiDAR measurements. These methods help the policy perceive the environment more comprehensively and improve navigation robustness. However, they increase system complexity due to additional perception modules, sensor calibration, modality synchronization, and fusion mechanisms, thus leading to higher computational costs. Furthermore, errors in segmentation, depth estimation, or sensor fusion may also propagate to the policy, and the enlarged observation space can make training more difficult.

\begin{table*}[t]
\centering
\caption{Comparison of interaction modeling strategies.}\vspace{-3pt}
\label{tab:interaction_model}
\centering
\setlength\tabcolsep{5pt}
\begin{tabular}{l l l l l l}
\toprule
\specialrule{0.1em}{1pt}{1pt} 
\tabincell{l}{\textbf{Modeling Strategy}} 
&\tabincell{l}{\textbf{Main Idea}} 
&\tabincell{l}{\textbf{Representation}} 
&\tabincell{l}{\textbf{Modeled Relation}}
&\tabincell{l}{\textbf{{Strengths}}}
&\tabincell{l}{\textbf{Limitations}}\\
\toprule

\tabincell{l}{\textbf{Joint State-Based}} 
&\tabincell{l}{{Encodes interactions as}\\{joint states and infers}\\ {conflicts from relative}\\{kinematic information}} 
&\tabincell{l}{{Structured joint state}\\{with robot and nearby-}\\{agent states}} 
&\tabincell{l}{{Robot-agent}} 
&\tabincell{l}{{Simple, compact, and}\\{computationally efficient}}
&\tabincell{l}{{Limited ability to model}\\{complex interactions}}\vspace{0.5em}\\

\specialrule{0em}{2pt}{2pt}
\tabincell{l}{\textbf{Attention-Based}}
&\tabincell{l}{{Learns weights over}\\{agents or interaction}\\{features to focus on}\\{decision-relevant}\\{information}} 
&\tabincell{l}{{Weighted agent or}\\{interaction features}} 
&\tabincell{l}{{Robot-agent,}\\{human-human}} 
&\tabincell{l}{{Focuses on influential}\\{agents and reduces}\\{irrelevant crowd}\\{information}}
&\tabincell{l}{{Depends on reliable}\\{detection and may not}\\{explicitly encode}\\{structured relations}}\vspace{0.5em}\\

\specialrule{0em}{2pt}{2pt}
\tabincell{l}{\textbf{Graph-Based}} 
&\tabincell{l}{{Models the navigation}\\{scene as a graph and}\\{performs relational}\\{reasoning via message}\\{passing among entities}} 
&\tabincell{l}{{Graph with entities}\\{as nodes and inter-}\\{actions as edges}} 
&\tabincell{l}{{Relations among}\\{robots, humans,}\\{and static obstacle}} 
&\tabincell{l}{{Strong relational modeling}\\{ability and useful for dense}\\{or heterogeneous scenes}}
&\tabincell{l}{{Requires accurate graph}\\{construction and usually}\\{has higher training and}\\{computational complexity}}\\
\bottomrule
\specialrule{0.1em}{1pt}{1pt} 
\end{tabular}
\vspace{-1.0em}
\end{table*}

\subsection{Interaction-Aware Methods}
\label{subsec:interaction_aware}

Interaction-aware methods are useful for navigation in crowded environments, where the robot's decision is influenced not only by individual agent states but also by mutual interactions among agents. As such, these methods explicitly model interactions among agents within RL frameworks and incorporate such relational information into policy learning. Existing methods encode interactions in different forms and can be classified into joint state-based, attention-based, and graph-based methods. The key differences among these interaction modeling strategies are summarized in Table~\ref{tab:interaction_model}, while Table~\ref{tab:interaction_aware_table} compares representative interaction-aware methods.

\subsubsection{Joint State-Based Methods} 

Joint state-based methods model robot-agent interactions through structured joint representations. By encoding information such as relative positions, velocities, distances, headings, and agent sizes, these methods allow the learned policy to reason about potential conflicts.

Chen et al.~\cite{chen2017decentralized} learned a value network that estimates time-to-goal values from joint states. During execution, each agent performs a one-step lookahead over candidate velocity actions and selects the action that maximizes the estimated value, which combines goal-reaching rewards and collision penalties. In this process, nearby agents are assumed to maintain their current velocities over a short time horizon. This method was later extended in~\cite{Chen2017social} by augmenting the reward function with social-norm penalties, encouraging socially aware behaviors such as passing on the right and overtaking on the left. To reduce the reliance on online lookahead under simplified motion assumptions, Everett et al.~\cite{everett2018motion} replaced value-based action evaluation with actor-critic policy learning. The method learns a policy that maps the robot state and nearby-agent states to a probability distribution over discrete actions. To handle a variable number of nearby agents, it uses a Long Short-Term Memory (LSTM)~\cite{hochreiter1997long} network to encode nearby-agent states into a fixed-length representation.

Zhu et al.~\cite{Zhu2023} instead focused on improving interaction-aware learning by modeling agent heterogeneity in terms of shape and speed. The method uses Orientated Bounding Capsules (OBCs) rather than homogeneous circles to represent robots and nearby agents, and incorporates capsule-based distances and relative orientations into the joint-state representation. It further introduces a velocity-related collision-risk metric, where faster agents induce larger forward risk regions.

\textit{Strengths and Limitations:} Joint-state-based methods encode the robot and agents in a shared state space, allowing the policy to use their relative positions and velocities for collision avoidance and social navigation. However, they mainly model robot-agent interactions, without identifying influential agents or capturing how agent-agent interactions affect the robot in dense crowds. These limitations motivate later methods that capture relational dependencies among agents.

\begin{table*}[t]
\centering
\caption{Comparison of graph-based interaction modeling strategies.}\vspace{-3pt}
\label{tab:graph_interaction_model}
\centering
\setlength\tabcolsep{5pt}
\begin{tabular}{l l l l l l}
\toprule
\specialrule{0.1em}{1pt}{1pt} 
\tabincell{l}{\textbf{Graph Type}} 
&\tabincell{l}{\textbf{Main Idea}} 
&\tabincell{l}{\textbf{Node}} 
&\tabincell{l}{\textbf{Edge}}
&\tabincell{l}{\textbf{{Strengths}}}
&\tabincell{l}{\textbf{Limitations}}\\
\toprule

\tabincell{l}{\textbf{Spatial Graph}} 
&\tabincell{l}{{Models interactions among}\\{entities at current time step}} 
&\tabincell{l}{{Robots and humans}} 
&\tabincell{l}{{Encodes inter-agent}\\{spatial interactions}} 
&\tabincell{l}{{Simple and effective for}\\{current-frame relational}\\{reasoning}}
&\tabincell{l}{{Limited ability to}\\{capture temporal}\\{evolution and entity}\\{heterogeneity}}\vspace{0.5em}\\

\specialrule{0em}{2pt}{2pt}
\tabincell{l}{\textbf{Spatio-Temporal}\\{\textbf{Graph}}}
&\tabincell{l}{{Extends graph reasoning}\\{from spatial relations to}\\{temporal evolution}} 
&\tabincell{l}{{Robots and humans}} 
&\tabincell{l}{{Spatial edges encode}\\{inter-agent relations;}\\{temporal edges encode}\\{motion continuity}} 
&\tabincell{l}{{Captures changing inter-}\\{actions and supports more}\\{anticipatory decisions}}
&\tabincell{l}{{Requires temporal}\\{history and increases}\\{model complexity}}\vspace{0.5em}\\

\specialrule{0em}{2pt}{2pt}
\tabincell{l}{\textbf{Heterogeneous}\\{\textbf{Graph}}} 
&\tabincell{l}{{Models different entity}\\{and relation types in}\\{one graph}} 
&\tabincell{l}{{Robots, humans, and}\\{static obstacles}}
&\tabincell{l}{{Encodes different types}\\{of spatial interactions}} 
&\tabincell{l}{{Suitable for cluttered}\\{and multi-entity scenes}}
&\tabincell{l}{{Requires accurate}\\{entity classification}\\{and more complex}\\{graph construction}}\\
\bottomrule
\specialrule{0.1em}{1pt}{1pt} 
\end{tabular}
\vspace{-1.0em}
\end{table*}

\subsubsection{Attention-Based Methods} Attention-based methods construct state representations in which nearby agents and interaction features are weighted according to their relevance to robot's navigation decision. In this way, the model can focus on agents that are closer to the robot, more likely to collide, or likely to affect path selection, while also emphasizing key interaction features, such as relative distance, relative velocity, heading direction, time-to-collision, and local crowd density.

Chen et al.~\cite{Chen2019} constructed latent pairwise interaction features for robot-human pairs using an MLP, with inputs including the robot's goal distance and velocity, and the human's relative position, velocity, radius, and distance to the robot. To capture human-human interactions, the method constructs a local map for each nearby human, where grid cells encode the occupancy and velocities of other humans around that person. This local map provides a coarse spatial representation of how surrounding humans may influence the motion of the current human. The resulting interaction features are fed into a self-attention-based pooling module, which assigns different importance weights to nearby humans and aggregates the weighted features into a compact crowd representation for action selection. This method was later extended in~\cite{LiuLucia2020} by processing humans and static obstacles through separate channels, enabling more appropriate navigation decisions.

While these methods improve spatial interaction modeling, they mainly reason about interactions at the current time step. Yang et al.~\cite{yang2023st} extended attention-based interaction modeling from single-frame spatial attention to spatio-temporal self-attention. The method uses the Transformer architecture~\cite{vaswani2017attention} to encode multi-frame robot-human joint states. A spatial Transformer encoder captures robot-human interactions, while a temporal Transformer encoder captures human motion trends and their changing relations with the robot over consecutive frames. The resulting spatio-temporal representation is used in a value-based RL framework for action selection.

\textit{Strengths and Limitations:} Attention-based methods improve interaction modeling by emphasizing decision-relevant agents, obstacles, or interaction features. However, they rely on accurate human detection and tracking, may miss structured relational dependencies among agents, and incur higher training and inference costs as the number of nearby agents or interaction features increases. Although they usually do not require manually labeled navigation labels, RL-based attention policies still require a large number of interaction samples, which are commonly collected from simulators or generated using demonstration policies for initialization.

\subsubsection{Graph-Based Methods} 
Graph-based methods model crowd navigation as relational reasoning over agents, obstacles, and their interactions. Nodes represent entities such as the robots, humans, or static obstacles, while edges describe their spatial relations, interaction strengths, or temporal connection. Compared with joint state-based or attention-based methods, graph structures provide a more explicit way to model robot-human, human-human, robot-robot, robot-obstacle, and group-level relations. According to the graph structure and relation modeling mechanism, existing methods can be grouped into spatial graph, spatio-temporal graph, and heterogeneous graph methods. The key differences among these graph modeling strategies are summarized in Table~\ref{tab:graph_interaction_model}.

\begin{table*}[t]{}
\footnotesize
\caption {Comparison of representative interaction-aware methods.}\label{tab:interaction_aware_table}\vspace{-3pt}
\centering
\setlength\tabcolsep{6pt}
\begin{tabular}{l l l l l l l} 
 \toprule
\specialrule{0.1em}{1pt}{1pt} 
\tabincell{c}{\textbf{Category}}
&\multicolumn{1}{l}{\textbf{Reference}}
&\tabincell{c}{\textbf{Year}}
&\tabincell{c}{\textbf{Main Idea}}
&\tabincell{c}{\textbf{Learned Model}}
&\tabincell{c}{{\textbf{Interaction Cue}}}  
&\tabincell{c}{\textbf{Modeled Relation}}\\ 
\toprule

\multirow{4}{*}{\tabincell{l}{\textbf{{Joint State-Based}}}}

& \tabincell{l}{\cite{everett2018motion}}      & 2018     &\tabincell{l}{{Learns a policy that maps}\\{joint states to a probability}\\{distribution over actions}} & \tabincell{l}{{LSTM-based actor-}\\{critic network}}  &\tabincell{l}{{Robot state and}\\{LSTM-encoded}\\{agent states}} &\tabincell{l}{{Robot-agent}}  \vspace{0.5em}\\

& \tabincell{l}{\cite{Zhu2023}}      & 2023     &\tabincell{l}{{Incorporates agents' shape}\\{and speed heterogeneity into}\\{joint state and collision risk}} & \tabincell{l}{{Heterogeneity-aware}\\{value network}}  &\tabincell{l}{{OBC-based}\\{distances and}\\{relative orientations}} &\tabincell{l}{{Robot-agent}}  \\

\specialrule{0.05em}{2pt}{2pt}
\multirow{5}{*}{\tabincell{l}{\textbf{{Attention-Based}}}}
& \tabincell{l}{\cite{Chen2019}}       & 2019     &\tabincell{l}{{Aggregates robot-human and}\\{human-human interaction}\\{features through self-attention}} & \tabincell{l}{{Self-attention pooling}\\{value network}}  &\tabincell{l}{{Pairwise robot-human}\\{features, human-}\\{centered crowd maps}} &\tabincell{l}{{Robot-human},\\{human-human}}  \vspace{0.5em}\\

& \tabincell{l}{\cite{yang2023st}}      & 2023     &\tabincell{l}{{Uses transformers to encode}\\{multi-frame robot-human}\\{interactions}} & \tabincell{l}{{Transformer}\\{value network}}  &\tabincell{l}{{Multi-frame robot-}\\{human joint states and}\\{spatio-temporal human}\\{motion cues}} &\tabincell{l}{{Spatio-temporal}\\{robot-human}\\{relation}}\\

\specialrule{0.05em}{2pt}{2pt}
\multirow{10}{*}{\tabincell{l}{\textbf{{Graph-Based}}}}
& \tabincell{l}{\cite{chen2020relational}}       & 2020     &\tabincell{l}{{Infers pairwise agent relations}\\{and propagates them through}\\{GNN for value estimation and}\\{human state prediction}} & \tabincell{l}{{Relational GNN}\\{with value esti-}\\{mation and motion}\\{prediction heads}}  &\tabincell{l}{{Inferred relation matrix}\\{from latent robot and}\\{human state features}} &\tabincell{l}{{Robot-human},\\{human-human}}  \vspace{0.5em}\\

& \tabincell{l}{\cite{liu2021decentralized}}      & 2021     &\tabincell{l}{{Converts a spatio-temporal}\\{crowd-navigation graph into}\\{separate RNN factors for}\\{policy learning}} & \tabincell{l}{{RNN-based actor-}\\{critic network}}  &\tabincell{l}{{Robot-human spatial}\\{edge features, robot}\\{temporal features,}\\{robot node features}} &\tabincell{l}{{Robot-human}\\{spatial relations,}\\{robot's motion}\\{continuity}}  \vspace{0.5em}\\

& \tabincell{l}{\cite{Liu2024}}      & 2024     &\tabincell{l}{{Models navigation scenes}\\{containing different types}\\{of obstacles and relations}\\{using heterogeneous graph}} & \tabincell{l}{{Heterogeneous}\\{GNN-based actor-}\\{critic network}}  &\tabincell{l}{{Multi-type edge}\\{features}} &\tabincell{l}{{Relations among}\\{robot, humans,}\\{static obstacles}}\\

\bottomrule
\specialrule{0.1em}{1pt}{1pt}%\vspace{2pt}
\end{tabular}
\vspace{-1.0em}
\end{table*}

a) \textit{Spatial Graph Methods:} These methods formulate the robot and humans as a spatial graph to model interactions. 

Chen et al.~\cite{chen2020relational} introduced relational graph learning for crowd navigation by constructing a directed graph, where each node represents an agent and each edge indicates an inferred relation between two agents. This relation can be interpreted as an influence weight, such as how strongly a nearby human affects the robot's decision or how one human influences another human's motion. Robot and human states are embedded into latent agent features using an MLP, from which a relation matrix is inferred through a pairwise similarity function. Based on this matrix, a GNN~\cite{kipf2016semi} propagates information among agents by aggregating neighboring node features according to the inferred relation weights. This message-passing process produces interaction-aware representations for both the robot and humans, which are used for value estimation and future human-state prediction, respectively. The learned value and prediction models are further used in multi-step lookahead planning for safe and efficient action selection.

Chen et al.~\cite{chen2020robot} further incorporated human attention into graph-based navigation. Instead of relying only on automatically inferred relations, the method learns attention weights from human gaze data and uses them to modulate the GNN adjacency matrix. The weighted graph allows the robot to aggregate crowd information according to the relative importance of nearby humans before value-based action selection. Jiang et al.~\cite{jiang2024learning} focused on edge-level asymmetric interactions and uses an edge-wise gating mechanism to control how much information is passed from each neighbor during message passing, enabling the policy to distinguish different influence levels in human-human and human-robot interactions.

b) \textit{Spatio-Temporal Graph Methods:} These methods capture interactions among agents across both spatial and temporal dimensions. By encoding how inter-agent relations evolve over time, they help the robot anticipate changing interaction patterns and make navigation decisions with greater foresight. 

Liu et al.~\cite{liu2021decentralized} developed a Decentralized Structural-Recurrent Neural Network (DS-RNN) based on S-RNN~\cite{jain2016structural}. The method constructs a spatio-temporal graph, where nodes represent agents, spatial edges encode human-robot relations at the same time step, and a 
temporal edge encodes the robot's motion continuity by connecting the robot node across two adjacent time steps. Separate RNN modules process spatial-edge, temporal-edge, and robot-node factors, and the resulting representation integrates spatial interactions, temporal motion information, and robot state for policy generation.

Unlike the above method, which models humans as independent individuals, Lu et al.~\cite{Lu2025} introduced group-aware spatio-temporal graph reasoning for crowd navigation. The method detects human groups and represents them as additional graph nodes, so that the graph contains the robot, obstacles, individual humans, and human groups. This allows the robot to reason about group-level social constraints, such as avoiding group intrusion, following or overtaking groups, and cooperatively passing oncoming groups. A spatio-temporal graph attention network estimates spatial relation strengths through attention, propagates information with GNNs, and uses LSTMs to capture the temporal evolution of these relations. The resulting spatio-temporal feature of the robot node is combined with the robot state for value-based action selection. 

c) \textit{Heterogeneous Graph Methods:} These methods model navigation scenes containing different types of entities (e.g., humans, robots, static obstacles) and their relations. By assigning type-specific node features and relation-specific message passing, these methods allow learned policies to distinguish dynamic-agent interactions, robot-robot interactions, and constraints imposed by static obstacles.

Liu et al.~\cite{Liu2024} modeled heterogeneous interactions between the robot and different types of obstacles using a heterogeneous spatial graph. Nodes represent the robot, humans, circular obstacles (e.g., trash bins), and line obstacles (e.g., walls), while edges encode interactions among these entities. A GNN aggregates heterogeneous node features using different transformations for different edge types, so that interactions with humans, circular obstacles, and line obstacles are encoded differently. The updated features are then fed into actor-critic networks to generate navigation actions. 

Zhou et al.~\cite{Zhou2025} extended heterogeneous graph reasoning to multi-robot crowd navigation. The method distinguishes the controlled robot, humans, and other robots, and constructs spatial graphs for different interaction types (e.g., human-robot, human-human, and robot-robot). A GNN propagates and aggregates information across these graphs, producing a controlled-robot embedding that encodes interactions with humans and other robots. This embedding is then used by a value-based network for action selection.

\begin{table*}[t]{}
\footnotesize
\caption {Comparison of representative hierarchical learning methods.}\label{tab:hierarchical_learning}\vspace{-3pt}
\centering
\setlength\tabcolsep{6pt}
\begin{tabular}{l l l l l l l} 
 \toprule
\specialrule{0.1em}{1pt}{1pt} 
\multicolumn{1}{l}{\textbf{Reference}}
&\tabincell{c}{\textbf{Year}}
&\tabincell{c}{\textbf{Main Idea}}
&\tabincell{c}{\textbf{Hierarchical}\\\textbf{Type}}
&\tabincell{c}{\textbf{High-Level}\\\textbf{Decision}}
&\tabincell{c}{\textbf{Low-Level}\\\textbf{Execution}}
&\tabincell{c}{\textbf{Inter-Level}\\\textbf{Coupling}}\\ 
\toprule

\tabincell{l}{\cite{jing2024two}}      & 2024    &\tabincell{l}{{Learns high-level subgoal}\\{selection and low-level control}\\{for long-range navigation}} &\tabincell{l}{{Subgoal-based}\\{hierarchy}} & \tabincell{l}{{Subgoal}}  &\tabincell{l}{{Control command}} &\tabincell{l}{{Top-down}}  \vspace{0.5em}\\

\tabincell{l}{\cite{chen2024environmental}}      & 2024 &\tabincell{l}{{Learns the high-level policy to}\\{adapt navigation modes according}\\{to the complexity of environment}} &\tabincell{l}{{Behavior-adaptive}\\{hierarchy}} & \tabincell{l}{{Behavior mode}}  &\tabincell{l}{{Control command}} &\tabincell{l}{{Top-down}}  \vspace{0.5em}\\

\tabincell{l}{\cite{Gao2024}}      & 2024  &\tabincell{l}{{Uses low-level collision or timeout}\\{failures as feedback for high-level}\\{subgoal learning}} &\tabincell{l}{{Feedback-enhanced}\\{hierarchy}} & \tabincell{l}{Subgoal}  &\tabincell{l}{{Control command}} &\tabincell{l}{{Bidirectional}}  \vspace{0.5em}\\

\tabincell{l}{\cite{Du2025}}      & 2025     &\tabincell{l}{{Uses hallway-level congestion}\\{reasoning to decide when and where}\\{to pass through constrained areas}} &\tabincell{l}{{Congestion-aware}\\{hierarchy}} & \tabincell{l}{{Hallway enter/wait}\\{decision}}  &\tabincell{l}{{Control command}} &\tabincell{l}{{Top-down}}  \vspace{0.5em}\\

\tabincell{l}{\cite{gao2025hierarchical}}      & 2025    &\tabincell{l}{{Uses environment congestion to}\\{adaptively update subgoals}} &\tabincell{l}{{Congestion-aware}\\{hierarchy}}  & \tabincell{l}{{Subgoal}}  &\tabincell{l}{{Control command}} &\tabincell{l}{{Top-down}} \\

\bottomrule
\specialrule{0.1em}{1pt}{1pt}%\vspace{2pt}
\end{tabular}
\vspace{-1.0em}
\end{table*}

\textit{Strengths and Limitations:} Graph-based methods explicitly model interactions among humans, robots, and obstacles, enabling the learned policy to reason about their mutual influences during navigation. By encoding different interaction types (e.g., robot-human, human-human, robot-obstacle, and robot-robot), these methods are suitable for dense crowds, cluttered environments, and multi-robot scenarios. However, their performance depends on accurate graph construction and reliable perception inputs, such as agent detection and tracking, category recognition, and group identification. More expressive graph models also increase training complexity, limiting robustness in unseen or rapidly changing scenarios.

\subsection{Hierarchical Learning Methods}
\label{subsec:hierarchical_learning}

Hierarchical learning methods decompose a learned navigation policy into high-level decision-making and low-level motion execution. The high-level policy provides intermediate guidance (e.g., subgoals), while the low-level policy converts this guidance into control commands. Table~\ref{tab:hierarchical_learning} summarizes the key features of representative hierarchical learning methods.

Existing methods differ in the form of guidance produced by the high-level policy. Jing et al.~\cite{jing2024two} adopted a subgoal-based hierarchy for long-range navigation. The high-level policy generates subgoals that guide the robot away from dense crowds while maintaining progress toward destination. The low-level policy then takes the subgoal as guidance and outputs control actions for collision avoidance and motion execution. Chen et al.~\cite{chen2024environmental} trained the high-level policy to adapt navigation modes (e.g., obstacle avoidance or goal pursuit) according to environmental complexity. The method characterizes the surrounding environment using two LiDAR-based metrics, namely the variation rate and entropy of environment structure, and uses them to regulate navigation modes. Du et al.~\cite{Du2025} built a hallway map based on the temporal arrival intents of nearby agents, allowing each agent to select a hallway by balancing travel distance and congestion. A learned congestion predictor then serves as a high-level policy that decides whether the agent should enter the selected hallway or wait. Okunevich et al.~\cite{Okunevich2025} used a high-level social module to assign a learned social value to the local trajectory induced by each low-level RL action. The final action is selected by considering the navigation value and the learned social value.

Other methods further improve how high-level decisions are trained or updated. Gao et al.~\cite{Gao2024} incorporated low-level execution feedback into high-level policy training. When the low-level policy fails to reach the selected subgoal due to collision or timeout, the high-level policy receives a penalty, encouraging it to select more feasible and reliable subgoals. Gao et al.~\cite{gao2025hierarchical} introduced a congestion-aware subgoal update mechanism, where local congestion is estimated from LiDAR observations and used to adapt the subgoal update threshold. Higher congestion prompts earlier subgoal updates in crowded regions, reducing lingering near the current subgoal and helping the robot escape crowded traps.

\textit{Strengths and Limitations:} Hierarchical learning methods use high-level guidance to improve the learning efficiency and decision quality of low-level policies. However, their performance depends on the quality of this guidance, as poor high-level decisions may mislead the low-level controller. Training and coordinating multiple policies also introduce additional design complexity, and mismatches between hierarchy levels may degrade navigation performance.

\section{Learning-Augmented Classical Planning Methods}
\label{sec:learningAugmented}
Learning-augmented classical planning methods retain classical planners as the primary decision-making modules for navigation, while using learned modules to provide auxiliary online information. In these methods, the learned module adapts planner parameters (e.g., objective-function weights and obstacle-clearance margins), provides search guidance, or predicts future obstacle motion. The classical planner then uses this learned information to generate the final trajectory or control command. According to the role of learning modules, these methods can be grouped into planner parameter learning, search guidance learning, and obstacle prediction methods. Table~\ref{tab:learningAugmented} summarizes the key features of representative learning-augmented classical planning methods.

\subsection{Planner Parameter Learning Methods}

Planner parameter learning methods adapt the numerical parameters of classical planners using learning techniques. In these methods, the learned module adaptively adjusts parameters, such as objective-function weights, forward simulation horizons, and obstacle-clearance margins. These updated parameters are then used by the classical planner to compute the final trajectory or control command.

Zhu et al.~\cite{Zhu2025} integrated RL with a classical trajectory optimization pipeline. A policy trained using the Twin Delayed Deep Deterministic Policy Gradient (TD3) algorithm~\cite{fujimoto2018addressing} adaptively adjusts objective weights and constraint-related parameters (e.g., safety distance threshold and maximum speed) based on LiDAR observations, robot states, previous actions, local goals, and coarse guidance information. The adapted parameters are then used by the optimization pipeline to compute the trajectory and corresponding control commands.

Han et al.~\cite{Han2025} trained an RL policy to learn when and how to deviate from the nominal MPC-based action for collision avoidance. The MPC module generates a nominal action for reference path tracking, while a human avoidance component uses a gated recurrent unit to encode human trajectories and multi-head attention to model human-human and robot-human interactions, generating a collision-avoidance action. A learned fusion module then predicts a blending coefficient that combines human-avoidance action with the MPC action to produce the control command.

Some methods adapt the parameters of local planner using RL. Chang et al.~\cite{chang2021reinforcement} used Q-learning~\cite{clifton2020q} to tune the weights of the DWA evaluation function and the forward simulation horizon according to the robot pose, goal position, and nearby-obstacle information. Dobrevski and Sko{\v{c}}aj~\cite{dobrevski2024dynamic} further trained a deep RL policy to predict the weights of the DWA evaluation function from a short history of LiDAR measurements, goal position, and the robot velocities. The use of recent observations allows the policy to implicitly capture obstacle motion in dynamic scenes. In both methods, the final control commands are generated by DWA. Chang et al.~\cite{chang2025oppa} used a transformer model~\cite{damanik2024lics} to predict the desired obstacle-clearance margin from LiDAR point clouds and the robot's local path. The predicted margin adapts obstacle clearance to the available free space along the planned path and updates parameters of the Timed Elastic Band (TEB)~\cite{rosmann2012trajectory}, including the minimum obstacle distance and costmap inflation radius.

\begin{table*}[t]{}
\footnotesize
\caption {Comparison of representative learning-augmented classical planning methods.}\label{tab:learningAugmented}\vspace{-3pt}
\centering
\setlength\tabcolsep{6pt}
\begin{tabular}{l l l l l l l} 
 \toprule
\specialrule{0.1em}{1pt}{1pt} 
\tabincell{c}{\textbf{Category}}
&\multicolumn{1}{l}{\textbf{Reference}}
&\tabincell{c}{\textbf{Year}}
&\tabincell{c}{\textbf{Main Idea}}
&\tabincell{c}{\textbf{Classical Module}}
&\tabincell{c}{{\textbf{Learned Output}}}  
&\tabincell{c}{\textbf{Final Output}}\\ 
\toprule

\multirow{11}{*}{\tabincell{l}{\textbf{{Planner}}\\\textbf{{Parameter}}\\\textbf{{Learning}}}}

& \tabincell{l}{\cite{chang2021reinforcement}}      & 2021     &\tabincell{l}{{Uses Q-learning to adaptively}\\{tune parameters of DWA}\\{evaluation function}} & \tabincell{l}{{DWA}}  &\tabincell{l}{{Evaluation-function}\\{weights, forward}\\{simulation horizon}} &\tabincell{l}{{Control command}}  \vspace{0.5em}\\

& \tabincell{l}{\cite{Zhu2025}}      & 2025     &\tabincell{l}{{Uses learned policy to adaptively}\\{adjust optimization
objectives}\\{based on environmental context}} & \tabincell{l}{{Trajectory}\\{optimization}\\{planner}}  &\tabincell{l}{{Objective-function}\\{weights, constraint-}\\{related parameters}} &\tabincell{l}{{Local trajectory}}  \vspace{0.5em}\\

& \tabincell{l}{\cite{Han2025}}      & 2025     &\tabincell{l}{{Learns blending coefficients to}\\{fuse human-avoidance actions}\\{with MPC path-tracking actions}} & \tabincell{l}{{MPC}}  &\tabincell{l}{{Action blending}\\{coefficients}} &\tabincell{l}{{Control command}}  \vspace{0.5em}\\

& \tabincell{l}{\cite{chang2025oppa}}      & 2025     &\tabincell{l}{{Uses a transformer-based model}\\{to estimate safety margins for}\\{TEB parameter adaptation}} & \tabincell{l}{{TEB}}  &\tabincell{l}{{Safety margin}} &\tabincell{l}{{Local trajectory}}  \\

\specialrule{0.05em}{2pt}{2pt}
\multirow{7.5}{*}{\tabincell{l}{\textbf{{Search}}\\\textbf{{Guidance}}\\\textbf{{Learning}}}}
& \tabincell{l}{\cite{chen2019horizon}}       & 2019     &\tabincell{l}{{Learns a sampling distribution}\\{from elite tree nodes to guide}\\{RRT*-based replanning}} & \tabincell{l}{{RRT*}}  &\tabincell{l}{{GMM-based}\\{sampling distribution}} &\tabincell{l}{{Global path}}  \vspace{0.5em}\\

& \tabincell{l}{\cite{yang2023rmrl}}      & 2023     &\tabincell{l}{{Uses a learned collision-risk}\\{evaluator to guide DWA}\\{candidate action scoring}} & \tabincell{l}{{DWA}}  &\tabincell{l}{{Collision risk}} &\tabincell{l}{{Control command}}\vspace{0.5em}\\

& \tabincell{l}{\cite{kathuria2025learning}}       & 2025     &\tabincell{l}{{Learns socially informed reward}\\{maps to generate ORCA-trackable}\\{reference trajectories}} & \tabincell{l}{{ORCA}}  &\tabincell{l}{{Reward map encoding}\\{scene navigability and}\\{implicit social preferences}} &\tabincell{l}{{Control command}}  \\

\specialrule{0.05em}{2pt}{2pt}
\multirow{8}{*}{\tabincell{l}{\textbf{{Obstacle}}\\\textbf{{Prediction}}}}
& \tabincell{l}{\cite{heuer2023proactive}}       & 2023     &\tabincell{l}{{Incorporates human trajectory}\\{predictions into MPC cost function}\\{for proactive collision avoidance}} & \tabincell{l}{{MPC}}  &\tabincell{l}{{Human trajectory}\\{predictions}} &\tabincell{l}{{Control command}}  \vspace{0.5em}\\

& \tabincell{l}{\cite{Lindemann2023}}      & 2023     &\tabincell{l}{{Uses conformal prediction to}\\{calibrate learned trajectory forecasts}\\{for probabilistically safe MPC}} & \tabincell{l}{{MPC}}  &\tabincell{l}{{Human trajectory}\\{predictions with}\\{conformal prediction regions}} &\tabincell{l}{{Control command}}\vspace{0.5em}\\

& \tabincell{l}{\cite{Samavi2025}}      & 2025     &\tabincell{l}{{Integrates diffusion-based joint human}\\{trajectory prediction into a bilevel}\\{MPC that jointly optimizes robot}\\{actions and human predictions}} & \tabincell{l}{{MPC}}  &\tabincell{l}{{Human trajectory}\\{predictions compatible}\\{with robot's plan}} &\tabincell{l}{{Control command}}\\

\bottomrule
\specialrule{0.1em}{1pt}{1pt}%\vspace{2pt}
\end{tabular}
\vspace{-1.0em}
\end{table*}

\textit{Strengths and Limitations:} Planner parameter learning methods improve the adaptability of classical planners to changing environments by adjusting their parameters online. However, their performance depends on whether the learned parameters generalize to unseen scenarios. Poor parameter adaptation may make the planner unsafe or overly conservative. In addition, these methods may still require careful reward design, bounded parameter ranges, and diverse training scenarios.

\subsection{Search Guidance Learning Methods}

Search guidance learning methods use learned modules to bias the search or plan evaluation process of classical planners. In these methods, the learned module produces sampling distributions, confidence maps, reward maps, or risk values. The classical planner then uses these learned quantities to guide sampling, path extraction, or trajectory evaluation, while preserving its original planning mechanism.

Chen et al.~\cite{chen2019horizon} learned an online sampling distribution to guide the expansion of an RRT*-based search tree~\cite{karaman2011sampling}. When the current path is invalidated by dynamic obstacles, the method reuses and repairs the existing search tree by pruning invalid nodes and edges while retaining reusable tree components. It then collects promising nodes into an elite set according to their cost-to-come and heuristic cost-to-go values, and fits a Gaussian mixture model~\cite{bishop2006pattern} to these elite samples. New samples are drawn from the learned distribution, together with goal-biased and random samples, to focus tree expansion on promising regions while maintaining sampling diversity.

Qin et al.~\cite{Qin2021} trained a CNN through supervised learning to predict a confidence map from a global path, LiDAR measurements, robot odometry, and human states. The confidence map encodes socially compliant path preferences and is converted into a cost map, from which an A*-based search extracts a local path for the low-level controller. Kathuria et al.~\cite{kathuria2025learning} used inverse RL to learn a reward map from few-shot expert demonstrations, encoding scene navigability and implicit social interaction preferences. During online deployment, a short-horizon reference trajectory is generated from the reward map and executed by an ORCA-based local controller. 

Yang et al.~\cite{yang2023rmrl} combined an RL-based risk evaluation network with DWA. The method first computes an occupancy grid map, where each grid value indicates the probability of human occupancy within a future time window. A risk evaluation network then estimates human risk from this map. The DWA simulates short-horizon trajectories for candidate velocity actions and scores them by combining the learned human risk with static-obstacle clearance. The robot executes the action with the highest score.

\textit{Strengths and Limitations:} Search guidance learning methods encode useful planning knowledge through sampling distributions, confidence maps, or reward maps, thereby biasing the search process toward promising regions. This improves the efficiency and adaptability of classical planners in dynamic environments. However, their performance depends heavily on the quality of the learned guidance. Inaccurate or outdated guidance may mislead the planner in rapidly changing environments. These methods often require representative training data or additional mechanisms to ensure that learned guidance does not compromise safety or completeness.

\begin{table*}[t]
\centering
\caption{Comparison of hybrid coupling strategies.}\vspace{-3pt}
\label{tab:hybridtype}
\centering
\setlength\tabcolsep{6pt}
\begin{tabular}{l l l l l l}
\toprule
\specialrule{0.1em}{1pt}{1pt} 
\tabincell{l}{\textbf{Hybrid Type}} 
&\tabincell{l}{\textbf{Main Idea}} 
&\tabincell{l}{\textbf{Coupling Structure}} 
&\tabincell{l}{\textbf{Operational Level}}
&\tabincell{l}{\textbf{{Classical Output}}}
&\tabincell{l}{\textbf{Learned Output}}\\
\toprule

\tabincell{l}{\textbf{Global-Local}\\\textbf{Planning}} 
&\tabincell{l}{{Uses a classical global}\\{planner for long-range}\\{guidance and a learned}\\{module for local execution}} 
&\tabincell{l}{{Hierarchical}} 
&\tabincell{l}{{Classical and learned}\\{modules operate at}\\{different levels}} 
&\tabincell{l}{{Global path}}
&\tabincell{l}{{Control command,}\\{local trajectory}}\vspace{0.5em}\\

\specialrule{0em}{2pt}{2pt}
\tabincell{l}{\textbf{Multi-Policy}\\\textbf{Adaptation}}
&\tabincell{l}{{Selects among learned and}\\{classical policies according}\\{to the navigation contexts}} 
&\tabincell{l}{{Parallel}} 
&\tabincell{l}{{Classical and learned}\\{modules operate at}\\{the same level}}  
&\tabincell{l}{{Control command,}\\{waypoint}}
&\tabincell{l}{{Control command,}\\{policy switching}\\{decision, waypoint}}\vspace{0.5em}\\

\specialrule{0em}{2pt}{2pt}
\tabincell{l}{\textbf{Hybrid Local}\\\textbf{Planning}} 
&\tabincell{l}{{Embeds learned intermediate}\\{quantities into local planning}\\{for final command generation}} 
&\tabincell{l}{{Cascaded}} 
&\tabincell{l}{{Classical and learned}\\{modules operate at}\\{the same level}}   
&\tabincell{l}{{Corrected, optimized,}\\{or dynamically feasible}\\{control command}}
&\tabincell{l}{{Nominal control}\\{command, velocity}\\{selection cue,}\\{intermediate reference}}\\
\bottomrule
\specialrule{0.1em}{1pt}{1pt} 
\end{tabular}
\vspace{-1.0em}
\end{table*}

\subsection{Obstacle Prediction Methods}

Obstacle prediction methods provide anticipatory information for proactive replanning in dynamic environments. Classical methods, such as Kalman filter~\cite{lu2024fapp} and particle filter~\cite{chen2025particle}, can predict obstacle states and uncertainty. However, they typically rely on assumptions about the underlying motion model and probability distribution, which may not hold in real-world applications. In contrast, learning-based methods~\cite{rasouli2019pie,salzmann2020trajectron,kothari2021human,knoedler2022improving} can predict complex motion patterns from data without specifying motion dynamics or distributional forms. 

Nishimura et al.~\cite{Nishimura2020} used learning-based human forecasts to induce a stochastic collision-cost distribution and optimized an entropic risk measure over this distribution within the MPC framework. This allows the controller to account for both the expected collision cost and its uncertainty. Heuer et al.~\cite{heuer2023proactive} added a penalty term to the MPC cost function to discourage candidate actions that would bring the robot close to predicted human positions. However, in interactive crowd navigation, human motion prediction and robot planning are inherently coupled, as the robot's planned motion may influence human responses, while predicted human motion affects the safety and efficiency of the robot plan. Samavi et al.~\cite{Samavi2025} addressed this issue by integrating human trajectory prediction with MPC-based motion planning. The method first uses a diffusion model~\cite{ho2020denoising} to generate joint trajectory prediction samples for all humans in the scene. These samples are then incorporated into a bilevel MPC formulation that jointly optimizes the robot plan and ORCA-refined human predictions, with the upper level optimizing robot actions and the lower-level ORCA problems enforcing collision-free predicted human motion with respect to other agents and static obstacles.

A remaining limitation is that many learning-based predictors do not explicitly provide calibrated uncertainty estimates, which may lead to unsafe or unnecessarily conservative planning decisions when the prediction is inaccurate. Lindemann et al.~\cite{Lindemann2023} addressed this issue by using conformal prediction~\cite{shafer2008tutorial} to quantify the uncertainty of learning-based trajectory predictions. The method constructs valid prediction regions around the predicted positions of dynamic obstacles using prediction errors measured on a calibration dataset. These regions are then incorporated into an MPC formulation. By planning trajectories that avoid these regions, the MPC can provide probabilistic safety guaranties while still exploiting learned trajectory forecasts. Strawn et al.~\cite{Strawn2023} extended conformal prediction to a RL-based planning framework by proposing a conformal predictive safety filter for pre-trained learning-based controllers. The method constructs conformal prediction regions around the predicted positions of dynamic agents. A safety filter is then trained to closely follow the nominal RL control sequence while modifying it when necessary to avoid the predicted agents and their uncertainty regions.

\textit{Strengths and Limitations:} Obstacle predictions enable planners to make anticipatory decisions by reasoning about future obstacle motions. However, their performance may degrade in highly dynamic or densely populated environments, where accurate prediction is hindered by limited sensing, high obstacle density, and computational constraints.

\section{Hybrid Planning Methods}
\label{sec:hybridplanning}

Hybrid planning methods integrate learned modules with classical planning in the online navigation pipeline. This distinguishes them from direct-policy learning methods, where learned models serve as the primary decision maker. Within direct-policy learning, hierarchical learning methods decompose navigation into high-level decision-making and low-level execution, but both levels are typically learned modules. In contrast, hybrid planning integrates learned modules with classical planners through hierarchical, parallel, or cascaded coupling structures. It also differs from learning-enhanced classical planning, where the classical planner remains the primary decision maker and learning mainly provides auxiliary information. In hybrid planning, learned and classical components jointly determine online navigation behavior.

In hybrid planning methods, learned modules may provide navigation actions, short-horizon trajectories, or intermediate references, while classical components provide global paths or control commands. According to the coupling strategy between learning and classical planning, existing methods can be grouped into global-local planning, multi-policy adaptation, and hybrid local planning methods. The key differences among these coupling strategies are summarized in Table~\ref{tab:hybridtype}, while Table~\ref{tab:hybridplanning} compares representative hybrid planning methods.

\subsection{Global-Local Planning Methods}

Global-local planning methods decompose navigation into a global guidance layer and a local execution layer. In these methods, classical planners are used as global planners to generate geometric paths towards the goal. Learned modules then operate at the local level, using onboard observations and global guidance to generate navigation actions or short-horizon trajectories. This design allows the robot to track the global path while adapting to nearby dynamic obstacles, collision-avoidance requirements, and kinodynamic constraints.

Wang et al.~\cite{wang2020mobile} used the A* algorithm~\cite{hart1968formal} to generate a global path and incorporated it into an RL-based local planner. The method first encodes local observations as image channels, while the global guidance is added as an additional guidance channel. These channels form the input at each time step. A sequence of such inputs over recent time steps is then fed into a Double Deep Q-Learning Network (DDQN)~\cite{van2016deep}. The network uses 3D CNN layers to extract spatial features and an LSTM layer to capture temporal information, and outputs the discrete motion actions. Guldenring et al.~\cite{guldenring2020learning} adopted a similar hybrid framework. A classical global planner provides waypoint guidance. An RL-based local planner is trained using PPO. The learned local planner takes LiDAR observations and upcoming waypoints as inputs, and outputs control commands to follow the global path while avoiding obstacle collisions.

Angulo et al.~\cite{angulo2023policy} proposed an RL-based steering function for classical planners (e.g., RRT~\cite{lavalle2001randomized}). Given LiDAR measurements, current and target state information, the learned policy generates kinodynamically feasible local trajectory to connect two states during planning, enabling the planner to account for non-holonomic constraints as well as static and dynamic obstacles. Xu et al.~\cite{Xu2024} combined RRT$^\mathrm{X}$ algorithm~\cite{otte2016rrtx} with a Q-learning-based local controller for kinodynamic motion replanning in dynamic environments. RRT$^\mathrm{X}$ algorithm maintains and repairs a global path as the environment changes, and decomposes the path into a set of boundary-value problems between consecutive waypoints. For each boundary-value problem, the Q-learning controller drives the robot toward the next waypoint under unknown system dynamics, external disturbances, and intermittent control updates.

\textit{Strengths and Limitations:} Global-local planning methods take advantage of both the long-range guidance and interpretability of classical global planners and the adaptability of learned models. However, their performance degrades when the global guidance becomes outdated in rapidly changing scenarios. Learned local modules may also deviate from the global guidance, and their effectiveness depends on policy generalization and compatibility with the global planner.

\begin{table*}[t]{}
\footnotesize
\caption {Comparison of representative hybrid planning methods.}\label{tab:hybridplanning}\vspace{-3pt}
\centering
\setlength\tabcolsep{6pt}
\begin{tabular}{l l l l l l l} 
 \toprule
\specialrule{0.1em}{1pt}{1pt} 
\tabincell{c}{\textbf{Category}}
&\multicolumn{1}{l}{\textbf{Reference}}
&\tabincell{c}{\textbf{Year}}
&\tabincell{c}{\textbf{Main Idea}}
&\tabincell{c}{\textbf{Classical Module}}
&\tabincell{c}{{\textbf{Classical Output}}}  
&\tabincell{c}{\textbf{Learned Output}}\\ 
\toprule

\multirow{12}{*}{\tabincell{l}{\textbf{{Global-}}\\\textbf{{Local}}\\\textbf{{Planning}}}}

& \tabincell{l}{\cite{wang2020mobile}}      & 2020     &\tabincell{l}{{Incorporates A*-based global}\\{guidance into a RL-based local}\\{planner for obstacle avoidance}} & \tabincell{l}{{A*}}  &\tabincell{l}{{Global path}} &\tabincell{l}{{Control command}}  \vspace{0.5em}\\

& \tabincell{l}{\cite{angulo2023policy}}      & 2023     &\tabincell{l}{{Learns an RL-based steering}\\{function to generate adaptive}\\{motion primitives for kino-}\\{dynamic replanning}} & \tabincell{l}{{RRT}}  &\tabincell{l}{{Target state pairs}\\{for local connection}} &\tabincell{l}{{Kinodynamically}\\{feasible local trajectory}}  \vspace{0.5em}\\

& \tabincell{l}{\cite{Xu2024}}      & 2024     &\tabincell{l}{{Uses RRT$^\mathrm{X}$ to maintain a}\\{collision-free global path and}\\{uses a Q-learning-based local}\\{controller to follow waypoint}\\{sequence under unknown}\\{dynamics and disturbances}} & \tabincell{l}{{RRT$^\mathrm{X}$}}  &\tabincell{l}{{Global path}} &\tabincell{l}{{Control command}} \\

\specialrule{0.05em}{2pt}{2pt}
\multirow{9}{*}{\tabincell{l}{\textbf{{Multi-}}\\\textbf{{Policy}}\\\textbf{{Adaptation}}}}
& \tabincell{l}{\cite{fan2020distributed}}       & 2020     &\tabincell{l}{{Switches among PID, RL, and}\\{conservative policies according}\\{to the scenario complexity}} & \tabincell{l}{{PID controller}}  &\tabincell{l}{{Control command}} &\tabincell{l}{{Control command}}  \vspace{0.5em}\\

& \tabincell{l}{\cite{wu2023risk}}      & 2023     &\tabincell{l}{{Learns navigation policies and}\\{risk-aware policy switching based}\\{on distributional-shift estimation}} & \tabincell{l}{{Safety fallback}\\{mechanism}}  &\tabincell{l}{{Safe action}} &\tabincell{l}{{Risk assessment,}\\{control command}}\vspace{0.5em}\\

& \tabincell{l}{\cite{hong2023obstacle}}       & 2023     &\tabincell{l}{{Uses CVAE-generated temporary}\\{waypoints for local dynamic}\\{obstacle avoidance and resumes}\\{navigation along the original path}} & \tabincell{l}{{A*}}  &\tabincell{l}{{Global path}} &\tabincell{l}{{Waypoint for dynamic}\\{obstacle avoidance}}  \\

\specialrule{0.05em}{2pt}{2pt}
\multirow{11}{*}{\tabincell{l}{\textbf{{Hybrid local}}\\\textbf{{Planning}}}}
& \tabincell{l}{\cite{sathyamoorthy2020frozone}}       & 2020     &\tabincell{l}{{Uses a model-based module to}\\{correct RL-generated velocities}\\{and avoid PFZs in human crowds}} & \tabincell{l}{{PFZ-based velocity}\\{correction module}}  &\tabincell{l}{{Corrected velocity}\\{away from PFZ}} &\tabincell{l}{{Nominal velocity for}\\{collision avoidance}}  \vspace{0.5em}\\

& \tabincell{l}{\cite{patel2021dwa}}       & 2021     &\tabincell{l}{{Uses DWA to construct a dynami-}\\{cally feasible velocity space and}\\{trains RL policy to select velocity}\\{for dynamic obstacle avoidance}} & \tabincell{l}{{DWA}}  &\tabincell{l}{{Dynamically feasible}\\{velocity candidates}} &\tabincell{l}{{Final velocity for}\\{collision avoidance}}  \vspace{0.5em}\\

& \tabincell{l}{\cite{LiuZhihao2024}}      & 2024     &\tabincell{l}{{Learns an escape velocity that}\\{encodes the robot's collision-}\\{avoidance responsibility and}\\{moving direction, and uses}\\{ORCA to compute final velocity}} & \tabincell{l}{{ORCA}}  &\tabincell{l}{{Goal-directed}\\{final velocity}} &\tabincell{l}{{Nominal velocity for}\\{collision avoidance}}\\

\bottomrule
\specialrule{0.1em}{1pt}{1pt}%\vspace{2pt}
\end{tabular}
\vspace{-1.0em}
\end{table*}

\subsection{Multi-Policy Adaptation Methods}

Multi-policy adaptation methods combine classical planning modules with learned modules at the same operational level. These modules may produce low-level commands, such as velocities, or higher-level decisions, such as intermediate waypoints. During execution, navigation behavior is adapted online by selecting between learning-based and classical planning-based policies according to the navigation context. 

Fan et al.~\cite{fan2020distributed} introduced a scenario-aware policy adaptation method that classifies navigation contexts into simple, complex, and emergent scenarios based on the distance to nearby obstacles and predefined distance thresholds. Thus, to maintain simplicity, the robot switches its policy depending on the scenario, such as a PID controller for simple scenarios, the RL-based policy for complex scenarios, and a conservative policy for emergent scenarios. Semnani et al.~\cite{semnani2020multi} adopted a similar framework, which selects control commands using an RL-based policy under normal scenarios, while switching to a classical method~\cite{semnani2020force} under high-risk situations.

Unlike the above methods that rely on hand-crafted policy switching rules, Wu et al.~\cite{wu2023risk} learned both the navigation policies and the risk-aware switching method, allowing policy selection to adapt to different navigation contexts. A flow-based density model~\cite{durkan2019neural} first estimates the probability density of latent feature-action pairs from the offline dataset. A higher density indicates that the corresponding state-action pair is more consistent with the training data. The density information is then used to train a Lyapunov density model (LDM)~\cite{kang2022lyapunov}, which predicts whether the future state-action trajectory will remain within the training-data distribution. During execution, the robot follows the learned policy when the LDM indicates low distributional-shift risk; otherwise, it switches to a safe action (e.g., emergency stop). Matsumoto et al.~\cite{matsumoto2024crowd} proposed a learning-based policy switching method using graph normalizing flows~\cite{liu2019graph}. The human states are encoded and passed through a graph normalizing flow to compute a likelihood-based switching score. During execution, the robot uses the learning-based navigation policy if the current human configuration has a high likelihood under the training distribution, and switches to ORCA-based policy otherwise. Zhu and Hayashibe~\cite{ZhuWei2023} proposed a learned safety supervisor, which decides whether to execute the learned policy or switch to a conservative maneuver for safe obstacle avoidance. 

Different from the above methods that switch among local control policies, Hong et al.~\cite{hong2023obstacle} proposed a waypoint adaptation strategy that modifies intermediate navigation goals. The method uses a Conditional Variational Autoencoder (CVAE)~\cite{doersch2016tutorial} to learn collision-avoidance waypoint distributions based on the static obstacles, human states, and robot's start and goal poses. During execution, if no human conflict is detected, the robot follows the waypoints provided by the classical global planner; otherwise, it navigates to a CVAE-generated waypoint for collision avoidance.

\textit{Strengths and Limitations:} Multi-policy adaptation methods combine the adaptability of learned policies with the reliability of classical planners, improving robustness across diverse navigation contexts and reducing reliance on a single policy. However, their performance depends on reliable scenario or risk assessment. Hand-crafted policy switching rules may not generalize well to different environments, while learning-based switching strategies require representative training data. Moreover, frequent or incorrect switching may lead to conservative, oscillatory, or inconsistent behaviors.

\subsection{Hybrid Local Planning Methods}

Hybrid local planning methods adopt a cascaded coupling between learned modules and classical local planners. In cascaded coupling, a learned module may first generate a nominal action, which is then corrected by a classical module to produce the final executable command. Alternatively, a classical module may first construct a dynamically feasible control command, from which the learned policy obtains the final command for collision avoidance. 

\begin{table*}[t]{}
\footnotesize
\caption {Comparison of representative training enhancement methods.}\label{tab:learningenhancement}\vspace{-3pt}
\centering
\setlength\tabcolsep{5.5pt}
\begin{tabular}{l l l l l l l} 
 \toprule
\specialrule{0.1em}{1pt}{1pt} 
\tabincell{c}{\textbf{Category}}
&\multicolumn{1}{l}{\textbf{Reference}}
&\tabincell{c}{\textbf{Year}}
&\tabincell{c}{\textbf{Main Idea}}
&\tabincell{c}{\textbf{Generalization Aspect}}
&\tabincell{c}{\textbf{Learning}\\\textbf{Formulation}}
&\tabincell{c}{\textbf{Enhancement}\\\textbf{Support}}\\ 
\toprule

\multirow{9}{*}{\tabincell{l}{\textbf{{Expert-Guided}}\\\textbf{{Training}}}}

& \tabincell{l}{\cite{tai2018socially}}      & 2018     &\tabincell{l}{{Uses BC to initialize a navi-}\\{gation policy from expert}\\{demonstrations and refines}\\{it through GAIL}} &\tabincell{l}{{Socially compliant navigation}\\{across common pedestrian-}\\{interaction scenarios}} &\tabincell{l}{{BC + GAIL}} & \tabincell{l}{{BC-pretrained}\\{policy from expert}\\{demonstrations}}   \vspace{0.5em}\\

& \tabincell{l}{\cite{xu2021human}}      & 2021     &\tabincell{l}{{Uses knowledge distillation}\\{to shape the reinforcement}\\{learning reward}} &\tabincell{l}{{Human-like collision avoidance}\\{in unseen multi-agent scenarios}} &\tabincell{l}{{Knowledge}\\{distillation}} & \tabincell{l}{{Expert policy-based}\\{reward shaping from}\\{human demonstrations}}   \vspace{0.5em} \\

& \tabincell{l}{\cite{qin2024non}}       & 2024     &\tabincell{l}{{Uses IL to initialize a navi-}\\{gation policy from expert}\\{demonstrations and refines}\\{it through TRL}} & \tabincell{l}{Non-homogeneous scenarios}  &\tabincell{l}{{IL + TRL}} & \tabincell{l}{{IL-pretrained}\\{policy from expert}\\{demonstrations}}  \\

\specialrule{0.05em}{2pt}{2pt}
\multirow{11}{*}{\tabincell{l}{\textbf{{Scenario}}\\\textbf{{Diversification }}}}
& \tabincell{l}{\cite{perez2021robot}}      & 2021     &\tabincell{l}{{Improves generalization by}\\{training on simple canonical}\\{indoor layouts that compose}\\{into more complex scenarios}} & \tabincell{l}{{Unseen complex indoor}\\{environments}}  &\tabincell{l}{{Compositional}\\{multi-layout}\\{learning strategy}} &\tabincell{l}{{Set of canonical}\\{layouts with common}\\{geometric patterns}}  \vspace{0.5em}\\

& \tabincell{l}{\cite{Sen2025}}       & 2025     &\tabincell{l}{{Improves generalization by}\\{training policy with diverse}\\{human behaviors generated}\\{via domain randomization}} & \tabincell{l}{{Diverse pedestrian behaviors}\\{and real-world human-robot}\\{interactions}}  &\tabincell{l}{{Domain}\\{randomization}} &\tabincell{l}{{Randomized ORCA}\\{pedestrian-behavior}\\{parameters}}  \vspace{0.5em}\\

& \tabincell{l}{\cite{Wu2025}}      & 2025     &\tabincell{l}{{Improves policy robustness}\\{by generating diverse yet}\\{realistic simulated human}\\{behaviors via diversity-}\\{aware crowd model}} & \tabincell{l}{{Unseen crowd scenarios}}  &\tabincell{l}{{RL-based diverse}\\{crowd generation}} &\tabincell{l}{{Diversity-aware}\\{crowd model}}\\

\bottomrule
\specialrule{0.1em}{1pt}{1pt}%\vspace{2pt}
\end{tabular}
\vspace{-1.0em}
\end{table*}

Sathyamoorthy et al.~\cite{sathyamoorthy2020frozone} used a model-based correction module to refine the nominal control commands generated by a learned policy, aiming to reduce the freezing robot problem. The method tracks humans, predicts their future positions, and constructs the Potential Freezing Zone (PFZ), which represents a conservative region where the robot may become stuck. Based on the learned control command, the model-based module applies an angular correction to avoid entering the freezing zone while maintaining progress toward the goal, and outputs the corrected velocity as the final navigation command.

Patel et al.~\cite{patel2021dwa} used DWA to construct the dynamically feasible velocity set and RL to select the final command from this set. The method first generates velocity candidates that satisfy the robot's acceleration and nonholonomic constraints. For each candidate velocity, it evaluates safety and goal-progress information over recent time steps, forming a compact observation that describes the quality of this candidate. Given this observation, the learned policy selects one feasible velocity pair for execution.

Brito et al.~\cite{Brito2021} integrated a learned intermediate-reference module into MPC-based local planning in dynamic environments. The learned policy encodes the robot state, nearby-agent states, and the global goal, and outputs a local position increment as an intermediate reference for MPC. This reference captures short-term interaction effects with surrounding agents, while MPC optimizes local control commands under kinodynamic and collision-avoidance constraints.

Liu et al.~\cite{LiuZhihao2024} integrated RL with ORCA to implicitly learn the robot's collision-avoidance responsibility and avoidance direction in crowds. Instead of assuming that each agent takes half of the avoidance responsibility, a PPO-trained policy infers an escape velocity from the robot state and nearby-human states. The magnitude of this escape velocity reflects the robot's avoidance responsibility, while its direction determines the avoidance direction. ORCA then uses the learned escape velocity to construct a collision-avoidance constraint and compute the final robot velocity.

Xu et al.~\cite{Xu_Nav2025} trained a PPO-based UAV navigation policy using separate representations for static and dynamic obstacles. Static obstacles are encoded as a voxel map, while dynamic obstacles are represented by bounding boxes, reducing the sim-to-real gap compared with raw image inputs. The learned policy outputs goal-directed velocity commands, which are checked by a VO-based safety shield and projected into safe velocity regions when they may lead to future collisions.

Srisuchinnawong et al.~\cite{Srisuchinnawong} introduced an online unsupervised learning module to augment classical local planners. The module correlates obstacle feedback with the local planner command and learns a proactive correction command at a higher update rate. The learned command is added to the planner command, allowing the robot to compensate for limited update rates of local planners and motion model mismatch while balancing smooth motion and collision avoidance.

\textit{Strengths and Limitations:} Hybrid local planning methods use cascaded coupling between learned modules and classical local planners to decouple planning requirements (e.g., dynamic feasibility, collision avoidance, and goal-directed motion). This allows each module to focus on the aspect it handles best, thereby improving the robustness, interpretability, and deployability. However, their performance depends on the reliability of learned intermediate quantities and may still be constrained by the assumptions, constraint formulations, or local minima of the underlying classical planner.

\section{Training Enhancement Methods}
\label{sec:learningenhancement}

Training enhancement methods improve learning-based navigation policies during the training process. These methods introduce expert guidance or diversify the training scenarios to reduce exploration difficulty and improve robustness and generalization. According to the type of training support, existing methods can be broadly divided into expert-guided training and scenario diversification methods. Table~\ref{tab:learningenhancement} summarizes the key features of representative training enhancement methods.

\subsection{Expert-Guided Training Methods}

Expert-guided training methods incorporate expert demonstrations into the training process of RL-based navigation policies to reduce exploration difficulty and impose useful navigation priors. 

Tai et al.~\cite{tai2018socially} used expert demonstrations to improve the training efficiency and social compliance of navigation policies. The method uses Behavior Cloning (BC)~\cite{torabi2018behavioral} to train an initial policy from expert trajectories. It then uses Generative Adversarial Imitation Learning (GAIL)~\cite{ho2016generative} to refine the policy by learning from entire trajectories and capturing how consecutive observations and actions evolve over time, thereby improving temporal consistency and social compliance. 
Xu and Karamouzas~\cite{xu2021human} incorporated human demonstrations as soft guidance through knowledge distillation~\cite{hinton2015distilling}. The method first derives expert policies from human demonstrations. Instead of cloning these expert policies, it applies knowledge distillation to shape the RL reward based on the deviation between the learned action and expert actions. This allows the policy to benefit from imperfect demonstrations without being strictly constrained by them. Qin et al.~\cite{qin2024non} used a model-based expert policy to improve the training efficiency and generalization of the low-level collision avoidance policy. The method first trains an Imitation Learning (IL)-based policy~\cite{yan2022mapless} to learn expert collision-avoidance actions from NH-ORCA~\cite{alonso2013optimal}. The initial policy is then fine-tuned through Transfer Reinforcement Learning (TRL)~\cite{zhu2023transfer} under non-homogeneous scenarios.

\textit{Strengths and Limitations:} Expert-guided policy learning helps improve training efficiency by providing useful prior behaviors. However, its performance depends on the quality and coverage of the expert information. Policies may also generalize poorly when deployed in scenarios that differ from the expert demonstrations or training environments.

\subsection{Scenario Diversification Methods}

Scenario diversification methods improve policy generalization by exposing learning-based navigation policies to diverse static obstacle layouts and human behaviors during training.

Perez et al.~\cite{perez2021robot} improved policy generalization to indoor environments with complex layouts through a compositional multi-layout training strategy. Instead of manually designing complex training scenarios, the method trains the policy on a set of simple layouts (e.g., corridors, doorways, and crossings). These layouts capture common geometric patterns in the environments, enabling the learned policy to generalize to unseen scenarios composed of similar structural elements.

Sen et al.~\cite{Sen2025} improved the generalization of human-aware robot navigation through domain randomization. The method randomizes ORCA parameters that control human behaviors, including human radius, velocity, and time horizon. These variations expose the policy to different personal-space requirements, walking speeds, and anticipatory avoidance behaviors, thereby improving its adaptability to diverse human behaviors in real-world environments.

Wu et al.~\cite{Wu2025} generated diverse human behaviors through a diversity-aware crowd model. The method assigns latent control codes to simulated human policies and trains a discriminator to infer these codes from trajectory features. By rewarding trajectories that are distinguishable across different codes, the model encourages diverse behaviors, such as different separation distances. A constraint is further imposed to keep these behaviors goal-directed and collision-aware rather than unrealistic or random.

\textit{Strengths and Limitations:} Scenario diversification improves the generalization of learned policies by expanding the training distribution over obstacle layouts and human behaviors. However, its effectiveness depends on how well the diversified scenarios cover real-world variations. Hand-crafted layouts or simulated crowd behaviors may still miss rare interactions, complex social norms, or unexpected human reactions.
\section{Conclusion and Future Directions} \label{sec:conclusions}

This paper presents a comprehensive review of representative works published primarily between 2015 and 2025. Classical planning methods are first revisited as algorithmic foundations, followed by a structured review of direct-policy learning, learning-augmented classical planning, hybrid planning, and training enhancement methods. For each category, this review summarizes the main problem settings, representative algorithms, key ideas, integration mechanisms, strengths, and limitations. Overall, recent studies show that learning is no longer limited to replacing classical planners with end-to-end policies. Learned models can serve as primary navigation policies, provide planner parameters, search guidance, obstacle predictions, intermediate references, policy-switching decisions, or training support. This diversity has expanded the design space of motion planning in dynamic environments, enabling better perception utilization, interaction modeling, online adaptation, and integration with classical planning structures. At the same time, existing methods still face challenges in generalization, safety verification, sim-to-real transfer, dense crowd navigation, prediction uncertainty, and robust learning-classical integration. 

The current challenges of learning-based methods motivate several future research directions, as discussed below.

\subsection{Sim-to-Real Transfer and Real-World Robustness}

Although learning-based motion planning methods have achieved promising simulation results, their deployment in real-world dynamic environments remains challenging. The sim-to-real gap arises from noisy sensor observations, localization and tracking errors, actuation delays, robot-model mismatch, and unmodeled environmental factors. This gap is further amplified by the random behavioral patterns of interacting agents. For example, pedestrians, vehicles, and robots may suddenly stop, change direction, yield, ignore the robot, or move in groups in ways that are difficult to reproduce in simulation. As a result, policies trained in simplified or idealized simulators may exhibit degraded performance, unsafe reactions, or overly conservative behaviors on physical robots.

Future research should develop learning-based planning systems that are explicitly designed for robust real-world deployment. Promising directions include high-fidelity simulators with realistic sensing, actuation, and crowd behaviors; domain randomization and domain adaptation to reduce distribution mismatch; real-world demonstrations and human-in-the-loop data collection to capture natural interaction patterns; and online adaptation or continual learning to improve policy robustness after deployment. In addition, learning-based planners should be combined with classical safety layers, such as MPC, CBFs, ORCA, DWA, or emergency-stop mechanisms, to handle unexpected failures during real-world execution.

\subsection{Safe and Certifiable Learning-Based Planning}

Safety verification remains a critical challenge for learning-based motion planning in dynamic environments. Direct-policy learning methods, especially RL-based policies, often use high-dimensional nonlinear models whose decision boundaries are difficult to interpret or formally verify. Their outputs can be sensitive to perception errors, prediction uncertainty, distribution shifts, and unseen interaction patterns. This issue becomes more challenging in dynamic environments, where moving obstacles and human agents continuously change the feasible motion space, and closed-loop interactions between the robot and surrounding agents can lead to unpredictable future states. As a result, empirical metrics such as success rate or collision rate are insufficient to fully characterize whether a learned planner can remain safe under rare, uncertain, or adversarial conditions.

Future research should move from empirical collision avoidance toward certifiable learning-based planning. One promising direction is to combine learned policies with formal safety mechanisms, such as MPC constraints, control barrier functions, reachability analysis, and conformal prediction. These mechanisms can provide safety envelopes or corrective actions when learned modules generate risky decisions. Another direction is to quantify and propagate uncertainty from perception, prediction, and policy outputs into the planning process, so that the robot can reason about risk rather than relying on deterministic predictions. Future systems should include failure detection and recovery mechanisms that identify when the learned module is operating outside its reliable domain.

\subsection{Dense Crowd and Multi-Agent Interaction Modeling}

Dense crowd navigation remains challenging for learning-based motion planning. In such environments, motion planning is not merely a multi-obstacle avoidance problem, but a closed-loop multi-agent interaction problem. The robot must reason about heterogeneous interactions among humans, robots, obstacles, and groups, while also considering social norms, occlusions, and uncertain human responses. Existing interaction-aware methods have improved relational reasoning through joint-state representations, attention mechanisms, and graph-based models. However, many methods still rely on pairwise interaction modeling or local agent-level features, which may be insufficient for representing crowd-level patterns such as group motion, lane formation, bottlenecks, queues, and opposing flows. In dense crowds, perception and tracking errors also become more frequent due to occlusions, and the computational cost of attention or graph-based reasoning can increase rapidly with the number of nearby agents.

Future research should develop scalable and socially aware interaction models for dense multi-agent environments. One direction is to move beyond individual-agent representations and learn crowd-level structures, such as density fields, flow patterns, group formations, and congestion regions. Another direction is to design more efficient graph and attention mechanisms, including sparse graphs, hierarchical graphs, agent clustering, and local-global attention, to support real-time planning in crowded scenes. In addition, learning-based planners should better couple interaction-aware prediction with motion planning, so that the robot can reason about how its own actions may influence surrounding agents. 

\subsection{Active Perception and Planning Coupling}

Many existing learning-based motion planning methods treat perception as an upstream module that provides inputs to the planner, such as detected obstacle states, predicted trajectories, risk maps, or semantic information. However, this sequential pipeline can be insufficient in dynamic environments. The robot's motion not only determines its future position, but also affects what it can observe and how surrounding agents may react. Occlusions, limited fields of view, tracking failures, and prediction uncertainty can cause the planner to make short-sighted or unsafe decisions if they are not explicitly considered during planning. Therefore, motion planning in dynamic environments should not only react to current observations and predicted trajectories, but also actively choose motions that improve future observability and reduce prediction uncertainty.

Future research should further couple active perception, prediction, and planning in a unified framework. One direction is to develop perception-aware planning methods that consider information gain, visibility improvement, tracking reliability, and occlusion reduction together with goal progress and collision avoidance. Another direction is to incorporate prediction uncertainty directly into planning costs, constraints, or risk measures, so that the robot can slow down, wait, or select more conservative paths when future agent motions are highly uncertain. In addition, prediction models should consider how surrounding agents may react to the robot's planned motions, since pedestrians, vehicles, and other robots may change their behaviors when the robot slows down, yields, detours, or approaches them. Belief-space planning, POMDP-based formulations, and learning-based active sensing policies may provide useful tools for reasoning under partial observability.

\subsection{Embodied AI for Semantic and Task-Level Planning}

Existing learning-based motion planning methods focus on mapping sensor observations, agent states, or interaction representations to navigation actions. However, real-world dynamic environments often require more than geometric collision avoidance and local interaction modeling. Robots must also understand scene semantics, task contexts, human intentions, and social norms. For example, a robot may need to distinguish between walkable and restricted areas, infer whether pedestrians are queuing or crossing, decide whether to wait or detour, and adapt its navigation behavior according to language instructions or task-level requirements. Embodied AI techniques, including vision-language models, large language models, world models, and foundation models, provide new opportunities to incorporate semantic understanding and high-level reasoning into motion planning.

Future research should integrate embodied AI with motion planning in a grounded and safety-aware manner. Rather than directly replacing low-level planners or controllers, embodied AI modules can provide semantic maps, task constraints, intermediate goals, social rules, candidate behaviors, or planner-switching decisions for classical planners, learned policies, or hybrid planning frameworks. Such integration may help robots connect language instructions, visual perception, human intention reasoning, and executable motion plans in dynamic environments. However, several challenges remain. Foundation models may produce hallucinated or poorly grounded outputs, and high-level semantic decisions must be translated into constraints, costs, or references that are feasible, verifiable, and compatible with real-time planning. Therefore, embodied AI should be coupled with reliable low-level planning, safety filters, and uncertainty-aware verification mechanisms. 

\balance
\bibliographystyle{ieeetr}
\bibliography{reference}

\end{document}